\documentclass[12pt]{article}

\usepackage{scicite}
\usepackage{times}
\usepackage{graphicx}
\usepackage{amsmath}
\usepackage{algorithm}
\usepackage{algorithmicx}
\usepackage{algpseudocode}
\usepackage{float}
\usepackage{caption}
\usepackage{hyperref}
\usepackage{bbm}
\usepackage[noabbrev,capitalise]{cleveref}

\title{Human-Level Reinforcement Learning through Theory-Based Modeling, Exploration, and Planning} 

\author
{Pedro A. Tsividis,$^{1,3\ast}$ Jo\~{a}o Loula,$^{1}$
Jake Burga,$^{1}$ \\
Nathan Foss$^{1}$, Andres Campero$^{1}$, Thomas Pouncy,$^{2}$ \\ Samuel J. Gershman,$^{2,3\diamond}$ Joshua B. Tenenbaum$^{1,3\diamond}$
\\
\\
\normalsize{$^{1}$Massachusetts Institute of Technology}\\
\normalsize{Cambridge, MA 02139, USA}\\
\normalsize{$^{2}$Harvard University, Cambridge, MA 02138 USA}\\
\normalsize{$^{3}$Center for Brains, Minds, and Machines}\\
\normalsize{Cambridge, MA 02139, USA}\\
\normalsize{$^{\diamond}$Equal contribution}
\\
\normalsize{$^\ast$To whom correspondence should be addressed; E-mail: tsividis@mit.edu.}
}

\date{}

\begin{document}

\baselineskip24pt

\maketitle

\textbf{Reinforcement learning (RL) studies how an agent comes to achieve reward in an environment through interactions over time. 
Recent advances in machine RL %
have %
surpassed human expertise at the world's oldest board games and many classic video games, 
but they require vast quantities of experience to learn successfully --- none of today's algorithms account for the human ability to learn so many different tasks, so quickly. Here we propose a new approach to this challenge based on a particularly strong form of model-based RL which we call Theory-Based Reinforcement Learning, because it uses human-like intuitive theories --- rich, abstract, causal models of physical objects, intentional agents, and their interactions --- to explore and model an environment, and plan effectively to achieve task goals. We instantiate the approach in a video game playing agent called EMPA (the Exploring, Modeling, and Planning Agent), which performs Bayesian inference to learn probabilistic generative models expressed as programs for a game-engine simulator, and runs internal simulations over these models to support efficient object-based, relational exploration and heuristic planning. EMPA closely matches human learning efficiency on a suite of 90 challenging Atari-style video games, learning new games in 
just minutes of game play and generalizing robustly to new game situations and new levels. %
The model also captures fine-grained structure in people's exploration trajectories and learning dynamics. 
Its design and behavior suggest a way forward for building more general human-like AI systems.
}

Games have long served as readily available microcosms through which to compare the flexibility and speed of human and machine learners. Atari-style video games are particularly revealing: While model-free RL systems inspired by basic animal learning processes \cite{sutton1981toward, schultz1997neural, watkins1992q, daw2006cortical} can learn to play many classic Atari games with a relatively simple neural policy network \cite{guo2014deep,mnih2015human,van2016deep,schaul2015prioritized,stadie2015incentivizing, mnih2016asynchronous,he2016learning,hessel2017rainbow}, humans learn much more quickly and generalize far more broadly \cite{tsividis2017a}. Human players can reach a competent level of play --- scoring much better than a random policy and making significant progress that generalizes to many game board variations --- in just a few minutes or a handful of episodes of play \cite{tsividis2017a}. Model-free RL systems require tens or even hundreds of hours of play to reach the same level \cite{van2016deep, hessel2017rainbow}. Even after reaching apparent mastery of a game, RL systems fail to generalize to even small variations of the game board or its dynamics \cite{kansky2017schema}.

Our goal here is to develop a computational account of how humans learn so efficiently and generalize so robustly, in terms that %
can also guide the development of 
more human-like learning in machines. In contrast to most work in RL and game AI  
\cite{silver2018general,vinyals2019alphastar, dota, kapturowski2018recurrent, jaderberg2018human,pmlr-v119-badia20a}, 
our intention is not to explain how a class of intelligent behaviors could evolve from scratch in a system that starts with zero knowledge. Rather we seek to capture learning as we see it in humans, modeling as closely as possible the trajectories people show from the beginning to the end of learning a complex task such as a new video game. To do this, we start from the knowledge that humans bring to these tasks: flexible but powerful inductive biases that are the product of evolution, development and culture, and that can be deployed to learn quickly even in environments very different from their prior experience at the level of pixels or raw observations.

Our approach can be seen as a particularly strong form of model-based RL \cite{sutton2018reinforcement}, 
which has a long history in both human and machine learning \cite{spelke1990principles,baillargeon2004infants,spelke2007core,csibra2008goal,gopnik2004theory,murphy1985role,carey1985conceptual,gopnik1997words, guestrin2003generalizing,pasula2004learning, pasula2007learning, xia2018learning, scholz2014physics, kansky2017schema,keramati2018strategic,oh2015action, fragkiadaki2015learning, chiappa2017recurrent, leibfried2016deep,ha2018world, racaniere2017imagination,kaiser2019model,watters2019cobra,schrittwieser2020mastering,hafner2021mastering}, but as of yet no proposals for how to capture human-level learning in complex tasks. In cognitive neuroscience, model-based RL has primarily been studied in simple laboratory tasks consisting of a small number of sequential decisions \cite{daw2011model, otto2013curse, kool2017cost}, 
where the algorithms and representations that are most successful do not directly transfer to the much more complex tasks posed by video games or the real world. In AI, model-based RL approaches are an active area of current neural network research and %
have the potential for greater sample-efficiency relative to %
model-free systems \cite{kaiser2019model,watters2019cobra, schrittwieser2020mastering,hafner2021mastering}. But these approaches do not attempt to capture human inductive biases or learn in human-like ways, and they are still far from achieving human-level learning trajectories in most games.

We refer to our approach as Theory-Based Reinforcement Learning, because it explicitly attempts to incorporate the ways in which humans,
from early childhood, are deeply guided by intuitive theories in how they learn and act --- how they explore the world, model its causal structure, and use these models to plan actions that achieve their goals \cite{murphy1985role,carey1985conceptual,gopnik1997words, lake2017building,allen2020}. Exploration stems from theory-based curiosity, an intrinsic motivation to know the causal relations at work in the world and to take actions directed to reveal those causes \cite{gopnik2004theory, xu2008intuitive, schulz2008going, cook2011science, Tsividis2014}. Causal model learning can be formalized as Bayesian inference of probabilistic generative models \cite{griffiths2009theory}, guided by priors based on intuitive theories --- especially core knowledge representations dating to infancy that carve the world into a natural ontology of objects, physics, agents, and goals \cite{spelke1990principles,baillargeon2004infants,spelke2007core,csibra2008goal}. Planning exploits intuitive theories to constrain the search over an otherwise intractably large space of possible action sequences; knowing which future states are worth thinking further about
helps even young children guide their search for effective actions and explanations \cite{SchulzError, tsividis2015hypothesis}.

We embody this general theory-based RL approach in a specific architecture that learns to solve new game tasks, which we call EMPA (the Exploring, Modeling, Planning Agent). We directly compare EMPA and alternative models on a new human-learning benchmark dataset (to be made available for download along with example videos of human and EMPA gameplay at \url{https://github.com/tsividis/Theory-based-RL} upon publication). The data consist of video game play from 300 human participants distributed across a set of 90 challenging games inspired by (and some directly drawn from) the General Video-Game AI (GVGAI) competition \cite{perez2014}. These games are simpler than classic Atari games but present fundamentally similar challenges. Some require fast reactive game play, while many require problem solving with nontrivial exploration and long-range planning. %
Like many real-world tasks and in contrast to Atari, a number of our games feature very sparse rewards, which present an additional challenge for artificial agents. As in Atari, people typically learn to play these games and generalize across levels in just a few minutes, or several hundred agent actions (Figure \ref{fig:EMPA_learning_curves_16}). %
By comparison, as we show below, conventional deep RL algorithms
typically take many thousands of steps to reach comparable performance on these games and often fail to solve even a single game level after hundreds of thousands of steps. EMPA aims to close this gap.

\begin{figure}%
\includegraphics[width=1.0\textwidth]{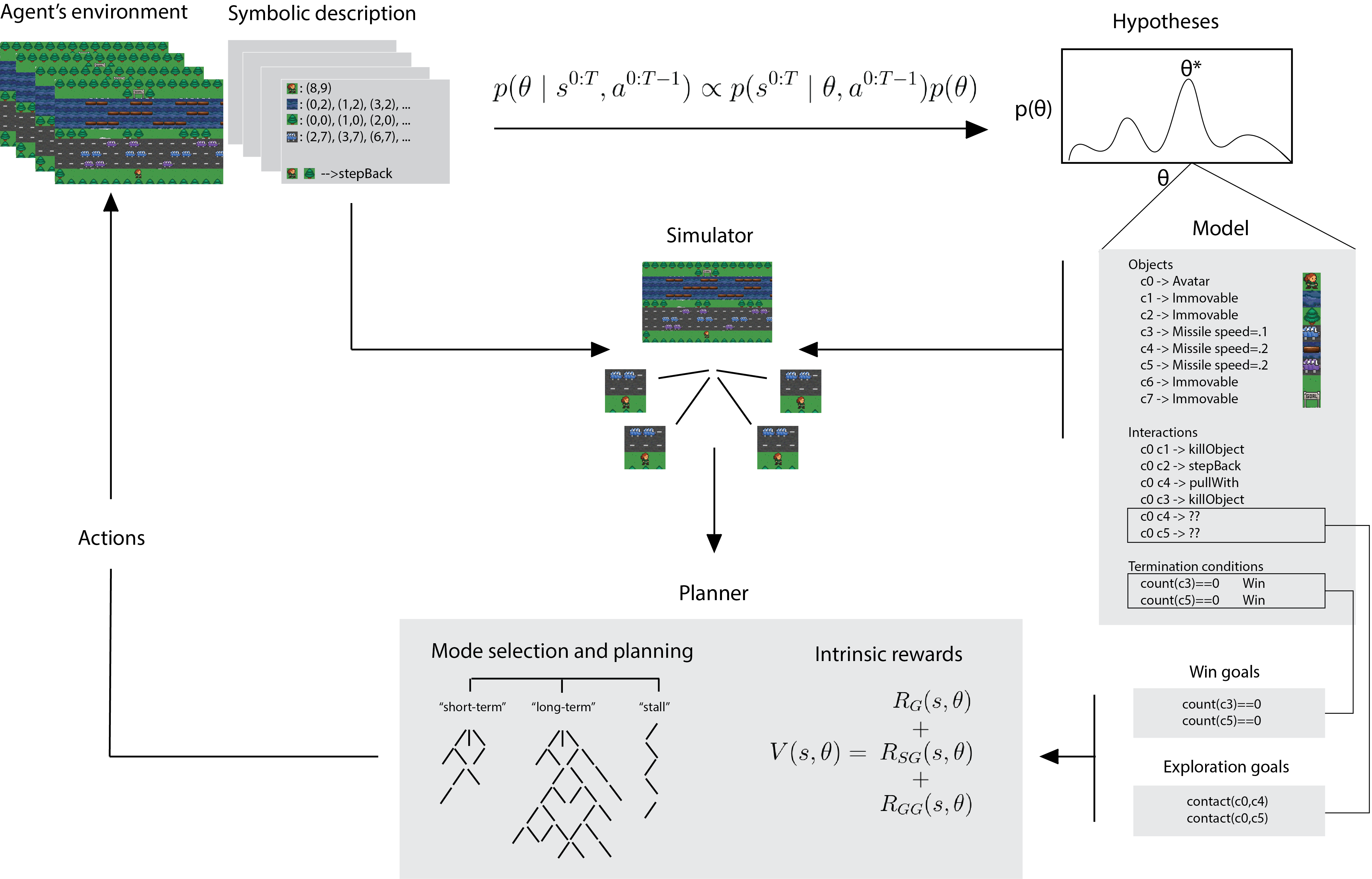}
\caption{\label{fig:EMPA_schematic}
Schematic representation of the EMPA architecture. EMPA takes as input a symbolic description of its environment, specified in terms of objects and their locations as well as interactions that occur between objects. Bayesian inference scores candidate models ($\theta$) on their ability to explain observed state sequences. Theory-based curiosity generates exploratory goals. The planner decomposes exploratory and win-related goals into subgoals and goal gradients, and uses this hierarchical decomposition to effectively find high-value actions for EMPA to take in the game environment.
}
\end{figure}

\begin{figure}%
\vspace{-2cm}
\hspace{-1.5cm}
\includegraphics[width=1.2\textwidth]{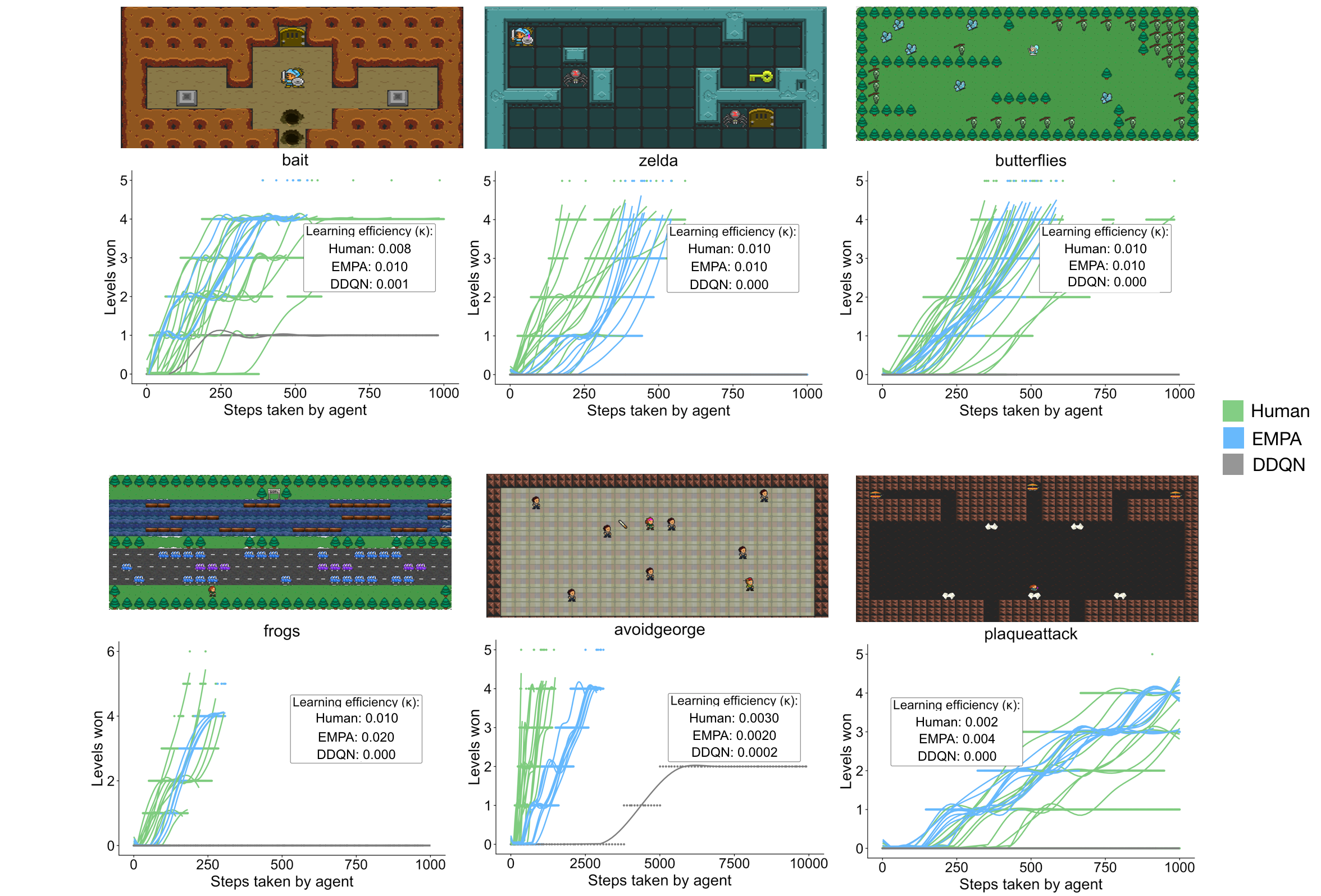}
\caption{\label{fig:EMPA_learning_curves_16} Humans (green) %
learn new games in a matter of minutes (corresponding to a few hundred steps). EMPA (blue) closely matches human learning curves in these and many other games, while DDQN (grey) generally makes little or no progress on the same time scale. Each dot shows the cumulative number of levels won at each point in time by an individual instance of each agent type (individual human or run of model); lines show smoothed scores from individual runs. Learning efficiency, $\kappa = \frac{\text{levels completed}}{\text{levels in game}} \times \frac{\text{levels completed}}{\text{steps to completion}}$, compactly summarizes each agent's performance over the indicated range of experience (10,000 game steps for ``Avoidgeorge'', 1,000 for the remaining games).
}\end{figure}

\vspace{\baselineskip}
\noindent {\bf Learning, exploration and planning}

\noindent
EMPA interacts with the environment in a continuous cycle of exploring, modeling, planning, and acting (Figure \ref{fig:EMPA_schematic}). 
The \emph {Modeling} component learns symbolic program-like descriptions of game dynamics, including types of entities, causal laws governing the interaction between entities, and win and loss conditions, as well as other sources of positive or negative rewards. Learned models support simulation of future states conditioned on a current (or imagined) game state and the agent's actions. The \emph{Planning} component searches efficiently for plans that achieve goals, guided by theory-based heuristics that decompose win and loss conditions into intrinsically rewarding reachable subgoals, and goal gradients that intrinsically reward steps toward those sub-goals. The \emph{Exploring} component generates theory-guided exploratory goals for the planner, so that the agent most efficiently generates the data needed to learn a game's causal interactions and win or loss conditions. Extended Data Figure 1
illustrates how these components allow the agent to learn dynamics models rapidly, even in games that pose severe exploration challenges, and to immediately generalize plans to win new game levels, even with substantially different object layouts than those encountered during initial learning. 

Implementing EMPA requires many detailed technical choices (see Methods and Supplementary Information), but here we focus on the key ideas behind each of the Exploration, Modeling and Planning components. 
The models learned by EMPA are object-oriented, relational, and compositional, and can be thought of as a projection of the core intuitive theories that allow infants to learn so rapidly about the real world, onto the virtual world of video games. Specifically, we represent models using a subset of the Video Game Description Language (VGDL)\cite{schaul2013video}, a lightweight language for Atari-like games in which all the GVGAI games are written.
(See examples in Extended Data Figures 2 and 3.) %
A VGDL description of a given game specifies the appearance and basic properties of all objects, which manifest as causal constraints on their dynamic properties: whether they move by default and in what directions, how quickly they move, what their goals are, and so on. Interaction rules specify the outcomes of contact events between pairs of objects as a function of their classes; these rules encompass natural concepts such as pushing, destroying, picking up, and so forth. Finally, termination conditions specify when a game is won or lost. These transition dynamics, together with a description of the game's initial state, completely specify a given game world. 

EMPA scores models according to a Bayesian criterion, with a hypothesis space corresponding to a restricted but still vast space of possible VGDL descriptions: for example, if we restrict to games with only $10$ unique classes of entities, there are over $3.6 \times 10^{37}$ possible VGDL descriptions or games that could be learned. 
The posterior probability of a specific game description $\theta$ conditioned on observed game states $s$ and actions $a$ is 
\begin{align}
p(\theta \mid s^{0:T}, a^{0:T-1}) \propto p(s^{0:T} \mid \theta, a^{0:T-1}) \, p(\theta).
\end{align}
The likelihood function, $p(s^{0:T} \mid \theta, a^{0:T-1})$, can be decomposed as 
\begin{align}
p(s^{0:T} \mid \theta, a^{0:T-1}) = p(s^{0}) \prod_{t = 1}^T p(s^{t}_{G} \mid \theta_{G}, s^{t}_{S}, s^{t}_{I}) p(s^{t}_{I} \mid \theta_{I}, s^{t-1}, a^{t-1}) p(s^{t}_{S} \mid \theta_{S}, s^{t-1}, a^{t-1})
\end{align}
which is a factorization of the state into goals, interaction events, and objects, respectively.

EMPA's exploration component implements a theory-based notion of curiosity that integrates its learning and planning functions, %
as the agent seeks to observe %
events that are most informative about unknown aspects of its hypothesis space. %
These exploratory goals can be formalized as a proxy reward function maximizing expected information gain \cite{lindley1956measure, bernardo1979expected}, but because of the nature of the models, the most informative events are always agent-object and object-object interactions for classes of objects that have not previously been observed in contact (see examples in Extended Data Figure 1). The agent thus explicitly makes plans whose goals are to produce these events. Crucially, this sense of curiosity is not myopic; EMPA often generates long-ranging plans whose sole purpose is to generate informative interactions. This enables EMPA to solve games that are highly challenging for traditional stochastic exploration methods.

EMPA's planner uses the maximum probability model as a simulator to imagine future states conditioned on candidate actions, searching a tree of future states to find high-value action sequences.
Sparse external rewards are a fundamental challenge for RL %
agents; EMPA addresses these by assigning several kinds of intrinsic rewards to imagined states. %
Treating exploration as a first-class goal, alongside winning, is one such intrinsic reward. EMPA also uses its domain theory to automatically decompose goals into easier-to-achieve subgoals, as well as goal gradients that help the planner more quickly find goals and subgoals by exploiting the spatial structure of the environment. A single intrinsic reward function %
summing these three quantities, $V(s, \theta) = R_{G}(s, \theta) + R_{SG}(s, \theta) + R_{GG}(s,\theta)$ (for goals, subgoals, and goal gradients, respectively) %
is applied to every environment, in all states $s$ the agent encounters or imagines, for any model $\theta$ it currently considers most likely. Generating intrinsic rewards in terms of these abstract goal-based concepts, which are in turn defined using the same %
description language that underlie EMPA's model-learning and exploration (see Methods), leads to a planner that effectively understands the agent's environments in a deep and generalizable way, and that is sufficiently powerful to find plans for most environments and tasks by simple best-first search through the space of imagined trajectories. The planner is also aided by Iterative Width (IW) pruning of insufficiently novel states from the search tree %
\cite{geffner2012width,lipovetzky2017best},
and several different search modes specialized for different time-scales of planning. A meta-controller manages decisions about whether to continue to execute previously-conceived plans or search for new ones, as well as which search mode to use, as a function of the highest probability model, the current game state, and the predictions of the previously-conceived plan. The meta-controller calls for long-term planning with deeper search trees %
when the environment is stable and predictable, short-term planning %
with shallower trees when the environment is rapidly changing or less predictable, %
and ``stall'' planning when other modes have failed to return satisfactory plans. %

\vspace{\baselineskip}
\noindent {\bf Evaluating EMPA and alternatives versus human learners}

\noindent We compared people, EMPA, and two leading single-agent deep-reinforcement learning agents, Double DQN (DDQN) \cite{van2016deep} and Rainbow \cite{hessel2017rainbow},
on a suite of 90 games (see Methods for human experimental procedures and a summary of the games; see Supplementary Information for detailed descriptions of all games). 
We also tested a number of ablated versions of EMPA, designed to highlight the relative contributions of its modules. %
Deep RL methods provide a valuable comparison point for our work as they represent the class of most successful and actively-developed RL agents on Atari-style tasks. We focus on DDQN in particular because it is a simple and widely-known variant of DQN, the original deep RL system for playing Atari-like video games; we also evaluate Rainbow, a recently developed method designed to combine the best features of DDQN and other DQN variants, and which achieves greater learning efficiency on many %
tasks. However, we hasten to add that we do not intend either of these deep RL methods to be seen as a baseline model for EMPA, or as a serious alternative account for how humans learn to play new games. 
Deep RL methods do not attempt to capture the background knowledge humans bring to a new task, nor how people explore and plan so efficiently using that knowledge, and so they necessarily will perform differently on our tasks. We explore the performance gap between humans and these methods only to quantify how much value there could be in building more human-like forms of RL --- our goal in developing EMPA and more generally the goal of the theory-based RL approach.

\begin{figure}%
\centering
\vspace*{-2.3cm}
\hspace*{-1cm}
\includegraphics[height=.9\textheight]{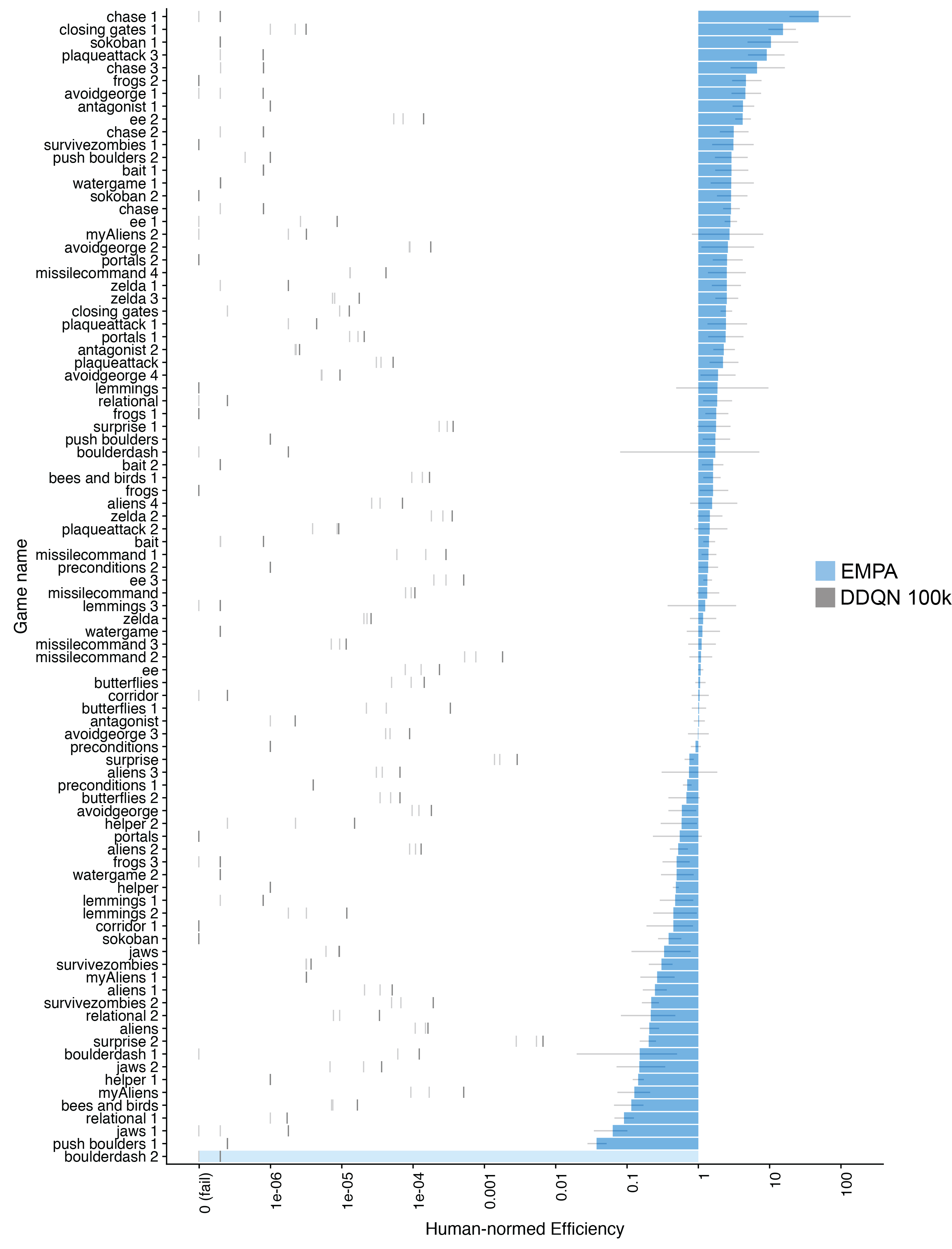}
\caption{\label{fig:composite_ratio}
Across 90 games, EMPA achieves comparable learning efficiency to humans, outperforming them on roughly two-thirds of games and underperforming them on roughly a third, but is almost always within an order of magnitude (0.1x-10x) of human performance. Shown in blue are mean scores and bootstrapped 95\% confidence intervals for 10 runs of EMPA on each game (adjusted in two games for sampling edge effects and excluding one game (light blue) on which no run of EMPA progressed beyond the first level; see Methods for details). By contrast, DDQN (grey vertical lines, with dark grey indicating the highest-performing run on each game) usually learns several orders of magnitude less efficiently than humans or EMPA do, and fails to win even a single level on multiple games.}
\end{figure}

\begin{figure}%
\centering
\includegraphics[width=.9\textwidth]{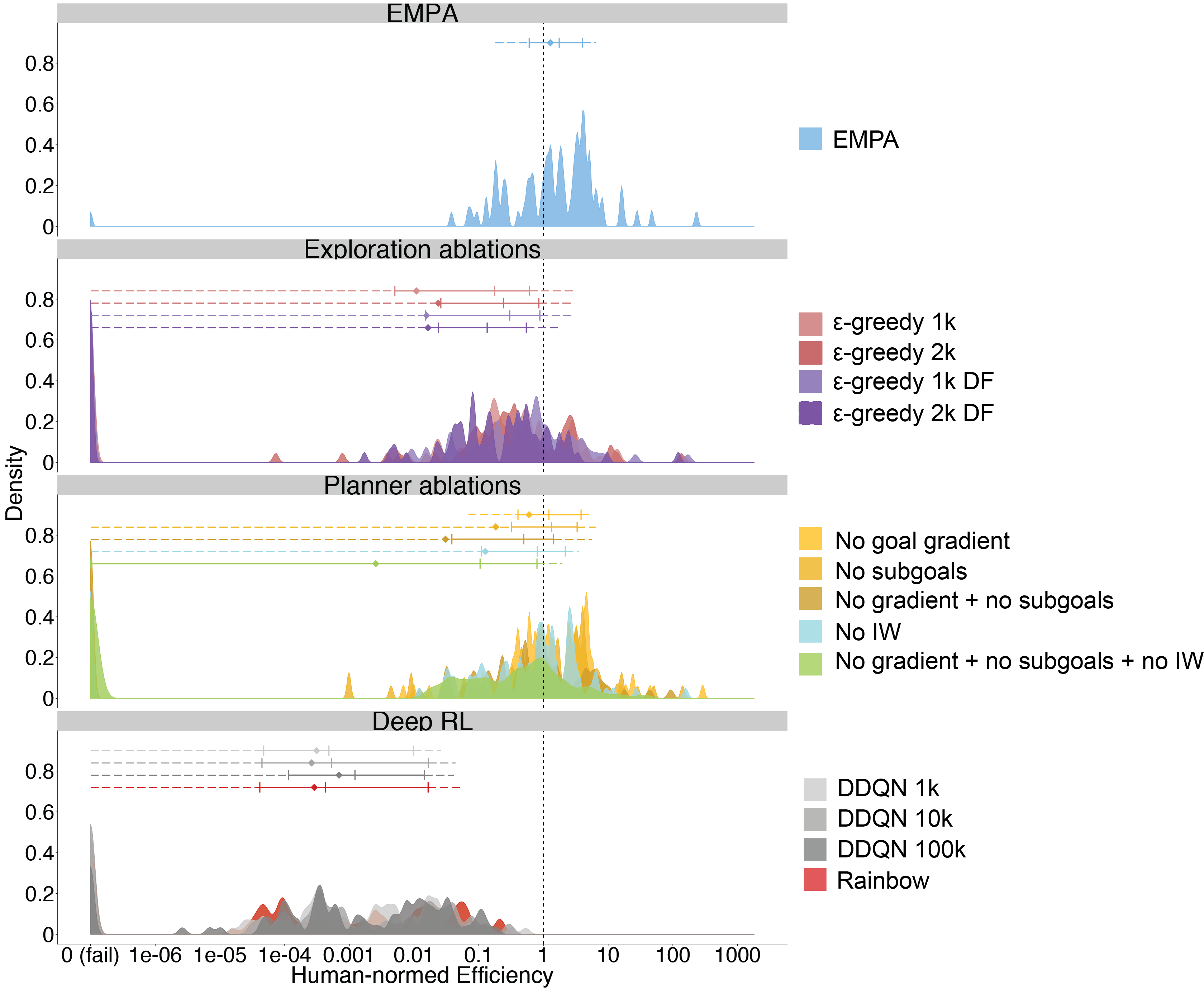}
\caption{\label{fig:ablations}
Ablations to EMPA's exploration (purple) or planning modules (yellow, turquoise, and green) cripple the agent's ability to win many games, producing comparable failure rates to DDQN, and suggesting that nontrivial exploration and planning abilities are a significant component of humans' ability to quickly perform well across many tasks. 
The vertical dotted line shows average human performance across all games. Horizontal line segments show $25$th-$75$th percentiles (with medians marked by a central hash); dashed lines show $10$th-$90$th percentiles; diamonds show geometric means.}
\end{figure}

Our primary criterion for comparing humans and models is learning efficiency, which reflects both depth and speed of learning: how many levels of a game does an agent win, and how quickly does it progress through these levels?
Learning curves show intuitively that EMPA matches human learning efficiency across most games (Figure \ref{fig:EMPA_learning_curves_16}; all 90 games shown in Extended Data Figures 5-7). To more quantitatively compare human and model performance, we define a learning efficiency metric, $\kappa = \frac{\text{levels completed}}{\text{levels in game}} \times \frac{\text{levels completed}}{\text{steps to completion}}$, a product of two terms measuring the degree to which a task is completed and the speed with which it is accomplished. On balance, EMPA matches human learning efficiency according to this metric: It is sometimes better and sometimes worse, %
but almost always (on 79 out of 90 games) within an order of magnitude (0.1x to 10x) of human performance %
(Figure \ref{fig:composite_ratio}). 
By contrast, DDQN almost always progresses much more slowly than humans: It is %
more than 100x less efficient on 67 out of 90 games, more than 1,000x worse on 45 out of 90 games, and over 10,000x worse on 22 out of 90 games. 
Rainbow, despite being significantly more sample-efficient than DDQN on Atari \cite{hessel2017rainbow}, performs no better on average %
here
(Figure \ref{fig:ablations}). This is likely %
in part due to the fact that humans generally complete our tasks within 1,000 actions, which is well within the margin during which both DDQN and Rainbow (and indeed all deep RL algorithms) are still exploring randomly. The fact that DDQN and Rainbow learn orders of magnitude more slowly than humans should thus not be surprising: They are not designed with the objective of learning %
as quickly as possible and do not attempt to embody %
the model-building or planning capacities that allow humans to learn %
without an extended period of stochastic exploration. %
But this comparison does illustrate starkly the value of building in more human-like cognitive capacities for RL --- potentially a thousand-fold gain in learning efficiency, which for this limited class of video game tasks, the EMPA model largely  achieves.

Although our game tasks are still relatively simple, their goals and dynamics vary considerably in ways that reflect the complexity and variability of tasks humans solve in the real world. For instance, in Bait (Figure 2, top left), the player must push boxes into holes in the ground in order to clear a path to a key that unlocks the exit. EMPA learns as fast as the median human, solving all five levels in fewer than 1000 steps; DDQN wins the first level as fast as humans do, but takes 250,000 steps to solve the first two levels and fails to solve the rest within the allotted 1 million steps. In Zelda (top middle), the player must also reach an exit that only opens once the player has obtained a key, but here the player has to complete this task while avoiding rapidly-moving dangerous creatures. Humans and EMPA can solve this game within 500 steps, whereas DDQN takes 100,000 steps to win four levels and fails to reach the fifth in 1 million steps. A variant of the game (Zelda 1) that requires the player to pick up three keys before reaching the door is roughly equally challenging for humans and EMPA, who can win all levels within 2,000 steps, but is far more challenging for DDQN --- it takes 800,000 steps to reach the third level and never progresses beyond it, despite the similarity of the variant to the original. In Butterflies (top right), the player must catch randomly-moving butterflies before they touch all the cocoons on the screen. EMPA performs as well as the best humans, winning all five levels within 1000 steps. DDQN solves the first four levels in as few as 20,000 steps, which is relatively fast compared to its performance on other games. However, it fails to solve the last level within 1 million steps, due to a subtle variation in the obstacle geometry which is fatal for the network's learned policy but poses no problem to EMPA's or humans' model-based planning.

The modular nature of EMPA allows us to ablate different components and thereby gain insight into their relative contributions to the full system's success across our suite of games (see Methods for details). Many games pose significant exploration challenges, requiring an agent to traverse a complex sequence of intermediate states (e.g., pick up a key to open a door, cross a road filled with dangerous moving cars, or push several boxes into holes) in order to reach a state where the game's win conditions can be discovered (Extended Data Figure 1). Such traversals are highly unlikely to occur by chance, and thus EMPA ablations that replace theory-based exploration strategies with $\epsilon$-greedy exploration --- essentially reducing exploration to stochastic exhaustive search --- fail %
dramatically in these games %
(Figure \ref{fig:ablations}). 
Likewise, ablations to any of EMPA's theory-based heuristics for planning (subgoal and goal gradient intrinsic rewards, or IW pruning) primarily lead to failures to discover solutions to even a single level for a subset of games %
(Figure \ref{fig:ablations}); ablating multiple planner components leads to higher failure rates and lower mean and median efficiency across games.
Many games challenge the ablated planners for the same underlying reason that they challenge exploration: When multiple subgoals must be reached in a specific sequence before any reward is received, successful plans are hard to discover by a simple model-predictive search.
Only EMPA's full complement of theory-guided planning heuristics and exploration mechanisms let it achieve human-level learning efficiency across the complete set of game environments tested.

\vspace{\baselineskip}
\noindent {\bf Fine-grained analysis of learning and exploring behavior}

\begin{figure}%
\centering
\vspace{-2cm}
\includegraphics[width=1\textwidth]{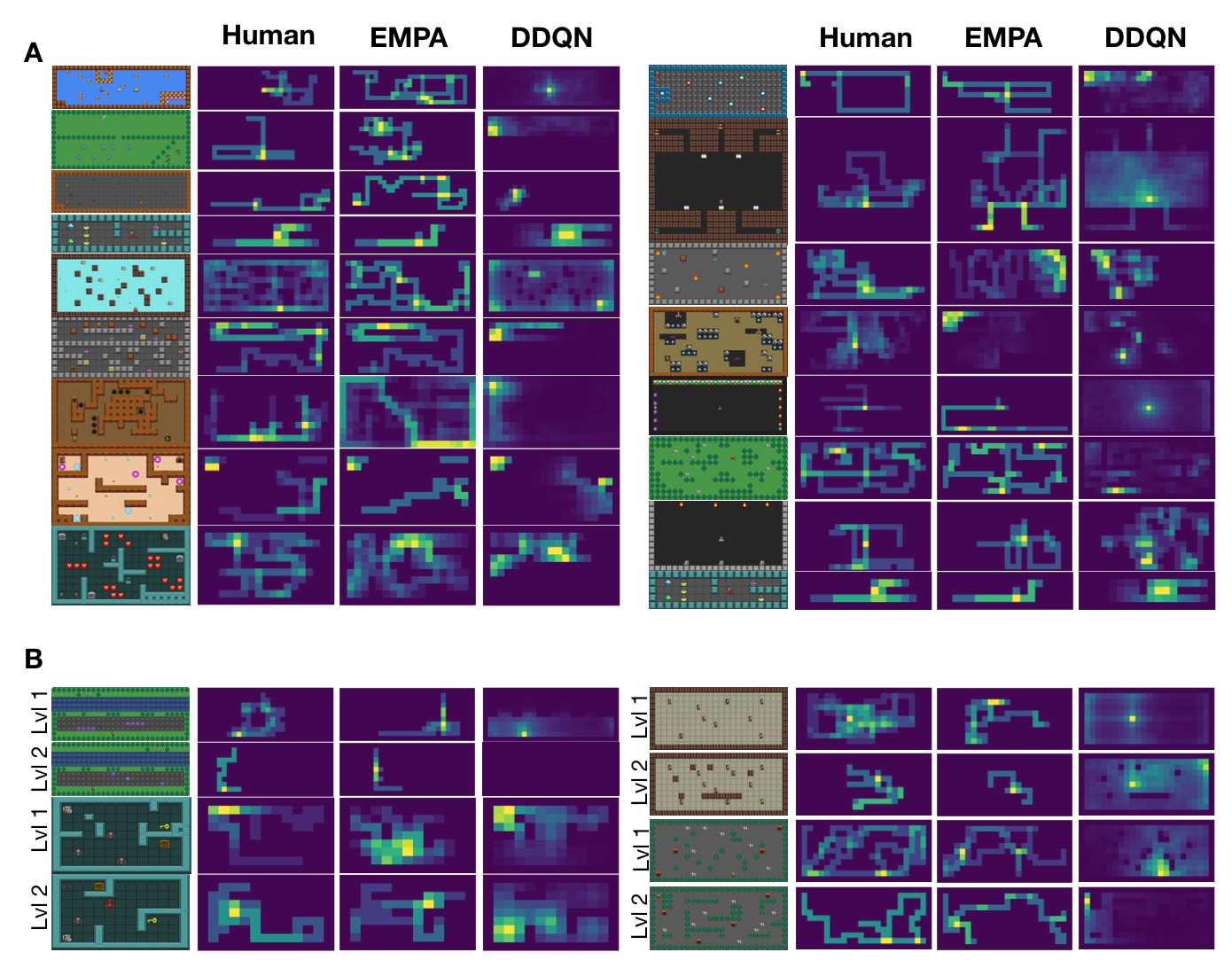}
\caption{\label{fig:grid} 
(A) Representative behavior traces of individual human and EMPA players show qualitative similarities across a wide range of games. %
Heat maps show the time spent by each agent in each location of the game board for a given level, across all episodes, normalized by that agent's total experience on that level. Yellow indicates the most-visited locations; dark purple locations are not visited. (B) Humans and EMPA sometimes produce more diffuse trajectories during initial levels of games, as they explore the results of interactions with most objects. An agent only needs to make contact with a few objects in order to win, resulting in more targeted behavior in subsequent levels, once dynamics are known.
}
\end{figure}

\noindent EMPA matches humans not only in its learning efficiency but also in the fine-grained structure of its behavior as it explores and solves game levels %
(Figure \ref{fig:grid}A)
. While the specifics of effective agent trajectories vary greatly across games as a function of their object dynamics and layouts, humans and EMPA share consistent similarities: They generate short, direct paths to specific objects, in both ``explore'' and ``exploit'' phases of learning a new task, %
reflecting efficient plans to reach objects with potential (epistemic) or known (instrumental) value. By contrast, DDQN's stochastic exploration and slow learning generate %
behavior that is much more diffuse and often restricted to a small subset of possible game board locations for long intervals (Figure \ref{fig:grid}A). %
Humans and EMPA also sometimes exhibit diffuse behavior early on in games, but then become more efficient on later levels %
(Figure \ref{fig:grid}B) as they move from exploring the properties of new objects to exploiting their learned models %
in order to win the game.

There are also significant differences between EMPA and humans, which in some cases lead to differences in behavior. EMPA is able to move faster than most humans do, and can thus find solutions that require fast coordination which humans often do not attempt. In addition, EMPA is always aware of all of its action choices, whereas humans sometimes neglect useful actions that would lead them to more efficient strategies. On the other hand, EMPA has weaker priors on higher-level aspects of games, leading it to sometimes be more exploratory than humans tend to be: for instance, EMPA doesn't know that games almost always have exactly one win condition, so after finding one way to win a game it may continue to explore new ways of winning, whereas humans tend to win more efficiently by exploiting a single already-known win condition. Empirically, these differences tend to balance out and produce similar qualitative and quantitative behavior, particularly when comparing EMPA and human performance to conventional deep RL agents.

\begin{figure}%
\centering
\includegraphics[width=1\textwidth]{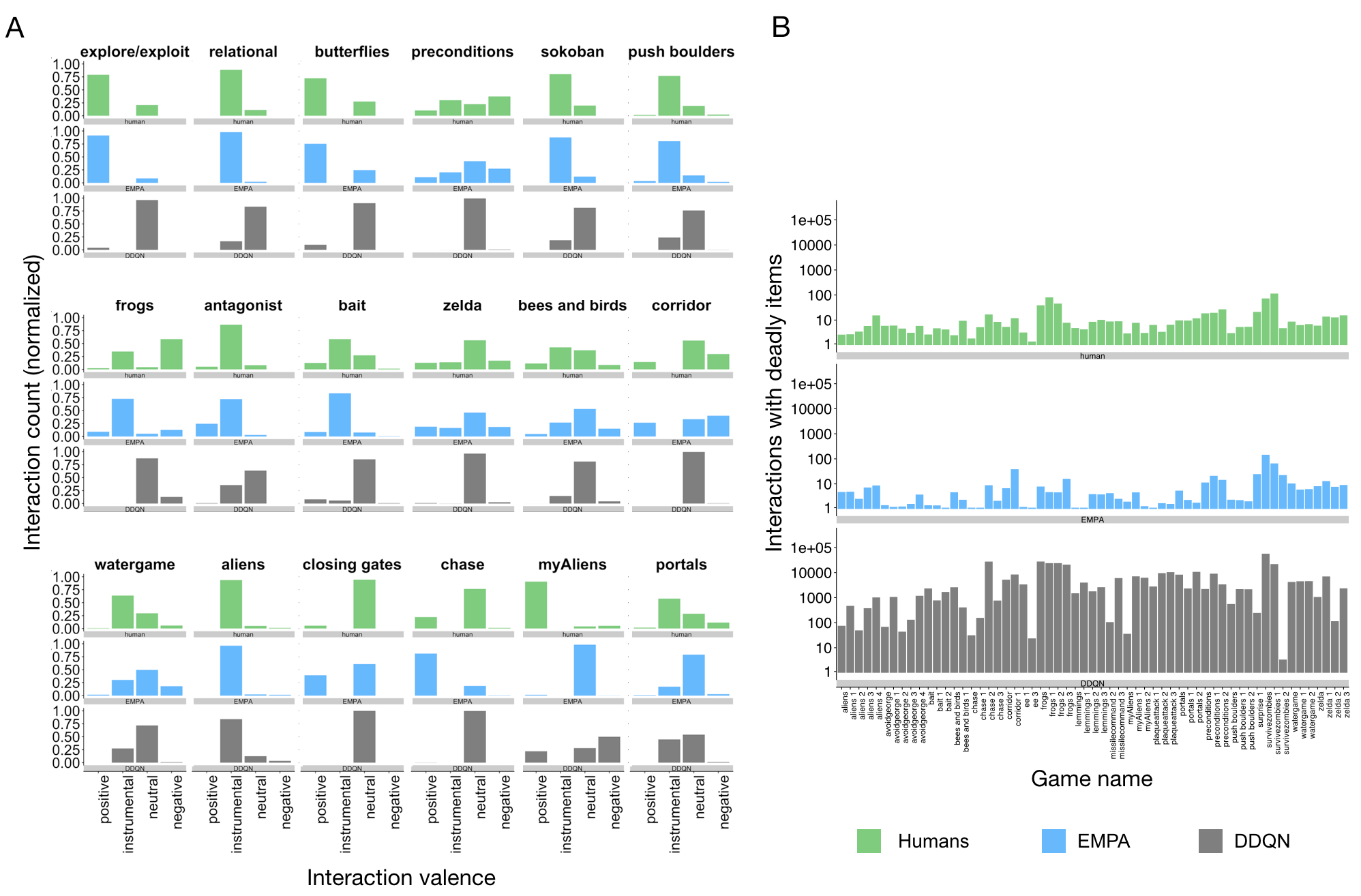}
\caption{\label{fig:behavioral_analysis} 
(A) EMPA and humans interact with objects of different types in similar proportions; shown here are distributions of agent interactions with objects classified as ``positive'', ``instrumental'', ``neutral'', and ``negative'', for 18 representative games (See Methods).
(B) Humans and EMPA learn quickly from interactions with items that result in immediate loss, and interact with such objects very few times, in absolute terms.}
\end{figure}

In order to quantify the ways in which EMPA explores, learns, and plans in human-like ways, we classified objects across 78 games as ``positive'', ``instrumental'', ``neutral'', or ``negative'' (see Methods) and coded how much time humans and model agents spent interacting with each class of objects. Across games, EMPA's profile of object interactions closely tracks the human profile (Figure \ref{fig:behavioral_analysis}A), 
matching the distribution of interactions better than DDQN does, particularly in slow-paced games in which only the agent moves (see Extended Data Figure 8). DDQN spends the greatest proportion of its time interacting with neutral objects (e.g., walls), as these are most numerous across all games and therefore will be collided with frequently as a result of DDQN's  random-exploration policy. By contrast, humans and EMPA display much more focused exploration that targets contact with object instances of unknown type; once these interactions are learned, further contact occurs only if an object is positive or instrumental (implicated in win-related goals, as in a key needed to open a door), or if contact is unavoidable given the constraints of the game state.

In the real world, humans (and many other animals) learn quickly when a class of objects is dangerous, and avoid them whenever possible. 
In our games, too, people typically interact with objects that result in immediate loss only a handful of times across all episodes (Figure \ref{fig:behavioral_analysis}B). EMPA behaves similarly, because of its rapid model-building and planning capacities: as soon as an object type is discovered to be dangerous, the planner %
avoids that object if at all possible. In contrast, 
DDQN typically interacts hundreds or thousands of times with deadly objects before it learns to avoid them 
(Figure \ref{fig:behavioral_analysis}B).

\vspace{\baselineskip}
\noindent {\bf Towards more human-like learning in AI}

\noindent A longstanding goal of both cognitive science and artificial intelligence has been to build models that capture not only what humans learn, but how they learn --- that can learn as efficiently and generalize as flexibly as people do. 
Our work represents a first step towards this goal, by achieving human-level learning efficiency in a large set of simple but challenging and widely varying video-game tasks.
Yet much remains open for future research. The assumptions of causal determinism and uniformity that let humans and EMPA learn so quickly --- assuming that interactions always produce the same effects, and objects of a given class have the same causal powers --- are powerful and pervasive. Even young children have these biases \cite{schulz2006god, gelman2003essential}. %
But while these principles often hold in the real world, humans can still learn when they are violated. Hierarchical Bayesian cognitive models capture this aspect of human flexibility \cite{griffiths2009theory, kemp2010learning}, and EMPA's Bayesian model learning could likewise be extended hierarchically to infer probabilistic interactions and object sub-types based on non-deterministic data. Humans also learn at least as much from others' experiences as from their own:
In the domain of video games, watching %
an experienced player or reading instructions %
supports even more efficient learning about game dynamics or strategy than playing on one's own \cite{tsividis2017a, tessler2017avoiding}. EMPA could be extended to capture at least basic forms of these social and linguistic learning abilities, by exploiting the interpretable structure of the learned model's dynamics and win-loss conditions. Most intriguingly, humans can use their learned models to achieve completely novel goals outside of those incentivized during training --- an important aspect of children's play more generally that shows up specifically in how children approach video games \cite{lake2017building}. For instance, a child might adopt the goal of losing (rather than winning) as quickly as possible, or of spending as much time as possible on a certain object without dying, or of teaching somebody else how the game works by touching each type of object exactly once. 
EMPA could be extended to play games in these ways as well, %
as its architecture makes it easy to specify arbitrary intrinsic goals for the planner.

In attempting to shed light on the inductive biases that make human learning so powerful, we were inspired by earlier model-based RL approaches based on Object-Oriented MDPs  \cite{diuk2008object}, Relational MDPs \cite{guestrin2003generalizing}, and probabilistic relational rules  \cite{pasula2004learning, pasula2007learning}. Our work more generally shares threads with recent methods that learn dynamics models giving objects, parts, and physics first-class status \cite{xia2018learning, scholz2014physics, kansky2017schema,keramati2018strategic}, and to a lesser degree, with methods that attempt more implicit unstructured model learning, including Bayesian RL \cite{oh2015action, fragkiadaki2015learning, chiappa2017recurrent, leibfried2016deep,ha2018world, racaniere2017imagination, ghavamzadeh2016bayesian}.
We go beyond such approaches by using explicit representations of physics, agents, events, and goals more directly inspired by human psychology \cite{spelke1990principles,baillargeon2004infants,spelke2007core,csibra2008goal,gopnik2004theory, xu2008intuitive, schulz2008going, cook2011science, Tsividis2014, SchulzError, tsividis2015hypothesis,pouncy2021model}, 
and by exploiting these inductive biases in each of the three main components of our agent --- exploration, modeling, and planning. 
But the inductive biases in EMPA represent only a step towards the full richness of human intuitive theories about physical objects, intentional agents, and how they can interact in the real world;
it remains an open question whether a theory-based approach would work as well in environments that do not so closely match the agent's hypothesis space of potential world models. Our hope is that by building an even richer compositional language for theories and combining this with more general-purpose program learning techniques \cite{ellis2018dreamcoder}, this approach can scale to increasingly varied and complex environments --- not just simple two-dimensional video games but more realistic three-dimensional environments, and eventually, the real world.

Despite our demonstration of the value of building in strong inductive biases, we do not mean to suggest that AI approaches with less built-in structure could not be developed to achieve similar performance. On the contrary, we hope that our work will inspire other AI researchers to set this degree of rapid learning and generalization as their target, and to explore how to incorporate --- whether through deep model-based learning \cite{kaiser2019model, watters2019cobra, schrittwieser2020mastering,hafner2021mastering}, 
meta-learning \cite{wang2018prefrontal, espeholt2018impala, jaderberg2018human,dasgupta2019}, simulated evolution \cite{salimans2017evolution}, or hybrid neuro-symbolic architectures \cite{battaglia2018relational, chang2016compositional, watters2017visual, zambaldi2018relational, garnelo2016towards} --- inductive biases like those we have built into our model. 
We suspect that any system that eventually matches human-level learning in games or any space of complex novel tasks will exhibit, or at least greatly benefit from, a decomposition of the problem into learning and planning, and from inductive biases, theory-based exploration, planning mechanisms like those of EMPA and humans. Where this prior knowledge ultimately comes from in human beings, and to what extent it can be acquired by machines learning purely from experience or is better built in by reverse-engineering what we learn from studying humans, remain outstanding questions for future work.

\section*{Acknowledgments}
The authors would like to acknowledge Aritro Biswas, Jacqueline Xu, Kevin Wu, Eric Wu, Zhenglong Zhou, Michael Janner, Kris Brewer, Hector Geffner, Marty Tenenbaum, and Alex Kell. The work was supported by the Center for Brains, Minds, and Machines (NSF STC award CCF-1231216), the Office of Naval Research (grant \#N00014-17-1-2984), the Honda Research Institute (Curious-Minded Machines), and gifts from Google and Microsoft.

\clearpage
\section*{Methods}

\subsection*{Informed consent}
Experiments were approved by the MIT Committee on the Use of Humans as Experimental Subjects and the Harvard Committee on the Use of Human Subjects. Informed consent was obtained from respondents using text approved by the committees.

\subsection*{Games and human behavioral procedures}

Games were generated from a core set of 27 games, 17 drawn from the General Video-Game AI (GVGAI) competition \cite{perez2014} and 10 in a similar style that we designed with particular learning challenges in mind. Each game is comprised of 4-6 levels, where different levels of a game use the same object types and dynamics but require different solutions due to their different layouts (board sizes and object configurations). For each of these 27 base games, we generated 1-4 variant games by altering different game features, such as the kinds of objects present, object dynamics, and the game's win or loss conditions. Humans found these variants as interesting and challenging to play as they did the core set of games (Extended Data Figure 4).
All 90 games are described in the Supplementary Information. %

300 human participants were recruited through Amazon Mechanical Turk and were paid \$3.50 plus a bonus of up to \$1.00 depending on their cumulative performance across all the games they played. Prior to playing the games, they were told only that they could use the arrow keys and the space bar, and that they should try to figure out how each game worked in order to play well.
They were also told that each game had several levels, and that they needed to complete a level in order to go to the next level, but that they could have the option to restart a given level if they were stuck, or to forfeit the rest of the game (and any potential earned bonus) if they failed to make progress after several minutes. Each participant played 6 games in random order. Twenty participants played each game, and no participant played more than one variant of a game. At the end of each game, participants were asked if they had ever played that game before; games for which participants indicated ``yes'' were excluded from analyses.

Humans interacted with all games via the arrow keys and the spacebar, and the games evolved through discrete time steps regardless of whether the human pressed a key or not. By contrast, all models interacted with an environment that evolved only after receiving a keypress, which for models additionally included the option of a ``no-op'' action. To make for a fair comparison between humans and models we use agent steps, rather than environment steps, as our metric when reporting learning curves and learning efficiency.

In order to test learning as it occurs independently of semantic content conveyed by object appearance, we displayed all games in ``color-only'' mode, where each object is depicted as a uniformly-sized block of a single color, and objects differ in appearance only in their color. Additionally, color assignments were randomized across participants to ensure that no consistent color-related semantic associations were accidentally conveyed to people. See Extended Data Figures 2-3 
for screen-shots of representative game levels in ``normal'' and  ``color-only'' mode. We did not ask participants to indicate whether they were color-blind, but no participants reported having trouble distinguishing colors in our games.

All models also played games in ``color-only'' mode. Results reported for EMPA were averaged over ten runs with different random seeds; EMPA ablations were averaged over five different runs, and results for DDQN and Rainbow were each averaged over three runs with different random seeds. 

All games can be played online in ``color-only'' mode, using a Chrome browser, at \\ \url{http://pedrotsividis.com/vgdl-games/}.

\subsection*{Model learning: representation}\label{model_representation}

EMPA builds models of each game environment using %
the Video Game Description Language, or VGDL \cite{schaul2013video}. %
VGDL specifies an environment in terms of three components corresponding to core aspects of human intuitive theories~\cite{Carey2009origin, lake2017building}:

\textbf{Objects:} A description of the appearance and basic properties of the different object classes in the game. These properties all manifest as causal constraints on the dynamic properties of the objects: whether they move by default, how and how often they move, whether they have simple goals (chasing or avoiding other objects), and if so, what those goals are. One object class is always the ``avatar'', representing and controlled by the player. We call the set of these properties the \emph{dynamic type} of each object. At every time-step, VGDL uses the dynamic type to calculate proposed positions for each object. If the proposed positions are empty, the objects move into those positions. If they do not, VGDL handles the resulting collisions by looking at the rules specified in the Interactions.

\textbf{Interactions:} A list of rules that specify how pairs of objects interact to produce events. These events correspond to intuitively natural action concepts, such as pushing, destroying, or picking up. All state changes beyond the movement patterns specified in the object-class descriptions are caused by these events. In turn, events are caused only by contact (collision) between objects.

For the sake of simplicity, EMPA only considers the following interaction types: \texttt{stepBack}, \texttt{bounceForward},  \texttt{reverseDirection}, \texttt{turnAround}, \texttt{wrapAround}, \texttt{pullWithIt}, \texttt{undoAll}, \texttt{collectResource}, \texttt{changeResource}, \texttt{changeScore}, \texttt{teleport}, \texttt{clone}, \texttt{destroy}, and \texttt{transform}, but could easily be extended to learn additional interaction types and thus model a far greater space of games.

\textbf{Termination conditions:} A specification of the win and loss conditions for a game. We restrict our scope to games that are won or lost when counts of particular objects on the screen reach zero (e.g., \texttt{WIN IF count(BLUE)==0}).

A VGDL description can also be thought of as procedurally specifying a Markov Decision Process consisting of the initial state of the game board, a transition function that specifies how the state evolves between successive time-steps (conditioned on the player's actions), and a reward function. The state at any time can be described by the object instances, their classes and locations; the avatar's internal state (the \texttt{agentState}), which tracks a set of resources the player has access to (e.g., health levels, or items carried); and any events occurring between pairs of objects that are participating in collisions at that time-step. It is useful to decompose this state $s$ into $s_{I}$, which refers to the pairs of objects that are participating in events as well as the nature of those events; $s_{S}$, which refers to the remaining objects; and $s_{G}$, which refers to the \texttt{Win}/\texttt{Loss}/\texttt{Continue} status of the environment. For the purposes of learning, we assume that the state-space is fully observed.

Our concrete task is to learn a distribution over the set of possible VGDL models, $\Theta$, that best explain an observed sequence of frames of game-play. For clarity, we decompose a particular model $\theta$ into three components corresponding to the objects (sometimes referred to as sprites) $\theta_{S}$, interactions, $\theta_{I}$, and termination conditions (or the agent's ultimate goals),  $\theta_{G}$.

The SpriteSet, $\theta_{S}$ consists of dynamic type definitions for each unique object class in the game. For a given class $c$, its definition is a vector $\theta_{S_{c}}$ assigning values to each of the parameters needed to fully describe a VGDL sprite (see \cite{schaul2013video} for these complete descriptions). 
One of these parameters (\texttt{vgdlType}) always encodes the abstract type of the object, which places high-level constraints on its behavior: whether it can move on its own or not, and if it can, whether it is an agent with chasing or avoidance goals, a random agent, or a projectile. The remaining parameters specify details of the object's behavior, such as its speed and orientation. The positions of sprites are updated at each step using simple programs that describe their behavior. For example:

\begin{itemize}
\item \texttt{nextPos(Missile, speed=2, orientation=Right, pos=(x,y)) = \\ (x+2, y)}
\item \texttt{nextPos(Random, speed=1, pos=(x,y)) = \\ random.choice([(x,y), (x+1,y), (x-1,y), (x,y+1), (x,y-1)]}
\end{itemize}

These programs imply a distribution over next positions for any given parameterization; we use this distribution to calculate per-object likelihoods in Equation \ref{eq:object_likelihood}.

The InteractionSet, $\theta_{I}$, consists of rules, one or more for each pair of object classes, specifying the effects of those classes' interactions. A rule $r_{i}$ is a tuple $(c_{j}, c_{k}, \xi, \pi)$ that prescribes a corresponding event, %
$e = event(r_i) = (c_{j}, c_{k}, \xi)$,
which is the application of predicate $\xi$ to an object of class $c_j$ whenever that object collides with an object of class $c_k$, and preconditions $\pi$ have been met. When an event $e$ occurs in the current state, we say that it is \texttt{True}. In accord with the intuitive concept of ``no action at a distance'', events are triggered only by contact between objects. %
For convenience, we denote by $pair(\cdot)$ the ordered pair of object classes implicated in a rule or event; a rule and its corresponding event always share the same class pair, $pair(event(r_i))=pair(r_{i})$. The predicates $\xi \in \Xi$ are observable state transformations on objects, such as \texttt{push}, \texttt{destroy}, \texttt{pickUp}, and so on. Preconditions in VGDL are restricted to statements of the form \texttt{count(agentState, item)}$<N$ or \texttt{count(agentState, item)}$>=N$, and are meant to capture common contingencies such as, ``The Avatar cannot go through a door unless it holds at least one key'' (or, ``at least three keys'', or ``has been touched by enemies no more than five times''). We use $\pi_{r_{i}}$ to refer to the precondition of rule $r_{i}$, and write $\pi_{r_{i}} = \texttt{True}$ when the precondition is met.

If an event $e$ occurs between a given class pair $(c_{j}, c_{k})$, all the other rules in the model that correspond to that class pair are expected to occur, provided their preconditions $\pi_{r}$ are met. We call these expected events \emph{implications}. Formally, the implications $\upsilon(e, \theta)$ of an event are: $\upsilon(e, \theta) = \{ event(r_{i}) \mid \big( pair(r_{i})=pair(e) \big) \land \big( \pi_{r_{i}}=\texttt{True} \big) \}$. Abusing notation, we will write that $\upsilon (e, \theta) = \texttt{True} \iff \upsilon_{j}=\texttt{True}, \forall \upsilon_{j} \in \upsilon (e, \theta) $.

Finally, we use $\eta(e, \theta)=\texttt{True}$ to denote that an event $e$, which is currently occurring, is \emph{explained} by a model, $\theta$. An event is explained in a model if there is a rule in the model that can account for the event. Formally, $\eta(e, \theta)=\texttt{True} \iff \exists r_{i} \in \theta_{I} \mid \big(event(r_{i})=e\big)\land \big( \pi_{r_{i}}=\texttt{True} \big)$.

The TerminationSet, $\theta_{G}$, consists of termination rules, $G$, each of which %
causes the episode to end (transition to a \texttt{Win} or \texttt{Loss} state) if some condition is met; if no termination conditions are met, the episode continues. For simplicity, EMPA only attempts to learn termination conditions that can be expressed in terms of the count of objects in a given class in some class equaling zero. This assumption can capture the win and loss conditions of almost all games (for instance, the agent loses when the number of avatars goes to zero, or wins when the number of goal flags goes to zero), and could easily be generalized if necessary. We use $\gamma(s_{I}, s_{S})$ to refer to the termination state $s_g$ prescribed by rule $g$, given the remaining aspects of the state, $(s_{I}, s_{S})$.

\subsection*{Model learning: Bayesian inference}

The posterior probability of a model after a sequence of time steps, $1:T$,  is $p(\theta \mid s^{0:T}, a^{0:T-1}) \propto p(s^{0:T} \mid \theta, a^{0:T-1}) \, p(\theta)$. For simplicity, we use a uniform prior over theories.

When an object is not involved in a collision, its movement patterns result only from its type and previous state. By contrast, when an object is involved in a collision, its subsequent state is a function only of the interaction rules that govern the classes of the objects involved in the collision. This enables us to decompose the likelihood function into two components corresponding to $s_{I}$ (the colliding objects and the events occurring between them) and $s_{S}$ (the freely moving objects). Once the state of all objects for a time-step has been resolved, the termination conditions can be evaluated, resulting in whether the the environment is in \texttt{Win}, \texttt{Loss}, or \texttt{Continue} status. This enables us to additionally factorize out the $s_{G}$ component of the state when we write the likelihood function, below.

The likelihood function can be decomposed as
\begin{align}
p(s^{0:T} \mid \theta, a^{0:T-1}) = p(s^{0}) \prod_{t = 1}^T p(s^{t}_{G} \mid \theta_{G}, s^{t}_{S}, s^{t}_{I}) p(s^{t}_{I} \mid \theta_{I}, s^{t-1}, a^{t-1}) p(s^{t}_{S} \mid \theta_{S}, s^{t-1}, a^{t-1})
\end{align}
where
\begin{align}\label{eq:goal_likelihood}
p(s^{t}_{G} \mid s^{t}_{S}, s^{t}_{I}, \theta_G) = \prod_{l=0}^{L} \mathbbm{1}[s^{t}_{G}=\gamma_{l}(s^{t}_{S}, s^{t}_{I})]
\end{align}
checks that the model's ensemble of termination rules correctly predict the  termination state at time $t$,
\begin{align}\label{eq:interaction_likelihood}
p(s^{t}_{I} \mid \theta_{I}, s^{t-1}, a^{t-1}) =
\begin{cases}
1-\epsilon & \big( \eta(e_{i}, \theta)=\texttt{True} \big) \land \big( \upsilon(e_{i}, \theta) = \texttt{True} \big), \forall e_{i} \in s^{t}_{I} \\
\epsilon & \text{otherwise}
\end{cases}
\end{align}
checks that all events in $s_t$ are explained by the rules and that all events expected to occur by the rules did, in fact, occur, and
\begin{align}\label{eq:object_likelihood}
p(s^{t}_{S} \mid \theta_{S}, s^{t-1}, a^{t-1}) = \prod_{k=0}^K p(o_{k}^{t} \mid \theta_{S}, o_{k}^{t-1}, a^{t-1})
\end{align}
encodes predictions for the objects $o_{1:K}$ not involved in collisions at time $t$. These per-object likelihoods are a function of the particular object parameterizations (dynamic types) specified by the model.

The above allows us to maintain a factorized posterior that corresponds to the three model components we aim to learn: the dynamic type for each class, the interaction rules that resolve collisions, and the termination rules that explain the win/loss criteria for a game. For the dynamic types we maintain an independent posterior over each parameterization, which we enumerate and update. For the interaction rules, we represent only the \emph{maximum a posteriori} hypothesis, which is easy to update because the interactions are deterministic. When we see a violation of a learned rule, we assume the violation is explained by some conditional rule instead, so we propose the minimal conditional rule that would explain it. The conditional rules we propose are limited to existential and universal quantifiers over aspects of the agent's state. For the termination rules we maintain a superset of possible explanations, which the planner tries to simultaneously satisfy.

To enable more efficient large-scale evaluation, we make several  simplifications to EMPA's learing module that could easily be relaxed if desired. We assume the agent has knowledge of which object corresponds to the avatar, its type (e.g., $\texttt{Shooter}$, $\texttt{MovingAvatar}$), as well as the projectile the avatar emits, if any. All of these model components could be learned by the inferential mechanisms above, at the cost of expanding the hypothesis space of dynamic types. Additionally, most games include a class of objects that function as ``walls'': they do not move, %
and they are %
more numerous than any other object class. We assume the agent knows the identity and dynamic type of walls, although their interaction rules (e.g., that they block the motion of other objects) must still be discovered. %
Given how numerous these objects are, and that none of them move, Bayesian evidence for their identity and dynamic type accumulates very rapidly across the first few game trials; hence EMPA's learning module could easily identify them, at the cost of more computation early on in learning a new game.

\subsection*{Exploration}

EMPA's exploration module works by setting epistemic goals for the planner, to observe the data most needed to resolve the learner's model uncertainty. Because the agent's knowledge is encoded in a posterior distribution over a hypothesis space of simulatable models, an optimal exploration strategy could search over agent trajectories and select ones that, over some time horizon, maximize expected information gain (EIG) \cite{lindley1956measure, bernardo1979expected} with respect to the agent's model posterior. 
However, the dynamic types of objects, $\theta_S$, can be quickly learned by observation and do not need to be explicit targets of exploration; only the interaction rules $\theta_{I}$ and termination conditions $\theta_{G}$ depend on the agent's actions.

Because of the object-oriented, relational nature of the model space, the uniform prior over interactions, the assumption that objects' interactions are determined completely by their classes, and the form of the likelihood function in Equation \ref{eq:interaction_likelihood}, seeing the events that occur after just one collision between objects of a given pair of classes $c_j, c_k$ is highly informative about that pair's interaction rules, for all models $\theta$ in the hypothesis space. The module therefore sets exploratory goals of generating interactions between every pair of classes whose interactions have not yet been observed. In addition, because interactions between the avatar and other objects can vary as a function of the \texttt{agentState}, exploratory goals between the avatar and other objects are reset when the \texttt{agentState} changes (See Extended Data Figure 1 for an example).
 
Because of the likelihood function for termination conditions in Equation \ref{eq:goal_likelihood}, the only states that are informative about termination rules for any hypothesis are ones where  \texttt{count($c_j$)==0} for some class $c_j$. When the model learns that the count of some class can be reduced, it sets reducing its count to zero as an exploratory goal that allows it to learn about $\theta_G$.

The initially large number of exploratory goals decreases as the model learns more about the game. At all points in the game, the planner mediates between exploratory goals and win-related goals as explained below.

\subsection*{Planning: overview}

The goal of the EMPA planner is to return high-value action sequences, together with the states predicted to obtain after each action. The EMPA planner takes as input a model, $\theta$ and state, $s$, and searches action sequences using the procedure detailed in Algorithm \ref{alg:planner_pseudocode}. Once a plan is found, it is executed until completion, or until the world state diverges sufficiently from the predicted state, in which case EMPA re-plans. We explain prediction-error monitoring and re-planning in a subsection below.

\begin{algorithm}
\caption{Pseudocode for the EMPA planner. The planner returns a high-value action sequence, as well as the states that are predicted to result from each action.}
\label{alg:planner_pseudocode}
\begin{algorithmic}[1]
\Function{plan}{$s_{0}$, maxNodes, plannerMode}
    \State
    numNodes, frontier, nodes = $0$, $[s_{0}]$,$[s_{0}]$
	\While{numNodes $<$ maxNodes}
        	\State 
        	\parbox[t]{\dimexpr\linewidth-\algorithmicindent}
        	{s = $\text{argmax}([v(s)$ for $s$ in frontier$])$\strut}
        	\State
        	frontier.remove(s)
            \For{$a \in$ actions}
            \State
            $s' \sim p(s' \mid s, a, \theta)$
            \State
            numNodes $+= 1$
            \If{novel($s'$, IW)}
            \State
            frontier.add($s'$)
            \State
            nodes.add($s'$)
            \EndIf
            \If{winCriterion($s'$, plannerMode)}
            \State
            \Return{actions($s'$.sequence), $s'$.sequence}
            \EndIf
            \EndFor
	    \State 
        $s'$ = argmax($[v(s)$ for $s$ in nodes$])$
    \EndWhile
    
\Return{actions($s'$.sequence), $s'$.sequence}
\EndFunction
\end{algorithmic}
\end{algorithm}

In order to plan efficiently in games with sparse rewards, %
EMPA evaluates plans in a hierarchical manner, with three distinct levels of representation: goals, subgoals, and goal gradients. 
 
\textbf{Goals} correspond to the fulfillment of known and hypothesized termination conditions, as well as to the fulfillment of contact goals specified by the exploration module. For example, if the agent's highest-probability theory  posits that the game is won by eliminating all objects of class $c_j$ (e.g., the goal is to pick up all the diamonds), the planner will set %
as a goal  \texttt{count}$(c_j) = 0$. %
Alternatively, before the agent has experienced any interactions with objects in $c_j$, the exploration module would set contact with any object of class $c_j$ as a goal: \texttt{contact(avatar, $c_j$)}. %

\textbf{Subgoals} represent partial progress towards termination goals. Since all termination goals specify conditions on the size of a set of objects in a certain class (of the form, $|c_{j}|=N$), subgoals are defined to be any change in the number of instances of a relevant class that moves in the desired direction relative to the current game state. For example, if the agent currently has a goal to pick up all diamonds, then any state which decreases the number of diamonds on the game board counts as fulfilling a subgoal.

\textbf{Goal gradients} represent preferences for states that are spatially closer to achieving a subgoal, computed based on the distances between pairs of objects in classes that are relevant to the agent's current goals. In our running example, a goal gradient expresses a preference for any action that moves the agent closer to the nearest diamond. But a goal gradient could also be defined based on bringing an object of one class closer to the nearest object of another class, if the agent has an exploratory goal to observe an interaction between those two classes.

Goals, subgoals and goal gradients are used both to define intrinsic rewards, supplementing the very sparse environmental reward structure, and to define the planner's completion criteria.

\subsection*{Planning: modes and metacontroller}\label{planning_modes}

The planner operates in three modes, ``long-term'', ``short-term'', and ``stall'', with a metacontroller that determines which mode to plan in at a given moment of game play (see Extended Data Figure 9).
Long-term mode returns the first plan that leads to a known or hypothesized win state, as well as any plan that satisfies the interaction goals of the exploration module. For a given game state, long-term mode searches up to a limit of 1,000 imagined states in order to find such a plan, and this limit progressively doubles over repeated failures to find successful plans. Short-term mode returns any plan that would have been returned by the long-term mode, as well as any plan that fulfills any subgoal. That is, if the number of objects in a class moves closer to the count specified by any termination condition, a subgoal has been reached and the planner returns that plan. In addition to terminating under different conditions, this mode searches shorter plans, initializing the search budget each time the short-term planner is called at one of three randomly chosen limits, $\{$200, 500, 1,000$\}$. %
Stall mode simply returns short plans that do not result in $\texttt{Loss}$ states, and searches only up to 50 nodes to find such plans; this mode is only run when the other two modes have failed to return a satisfactory plan. 

Because we did not attempt any performance engineering on the Python VGDL game engine, and this engine can slow down considerably when there are many objects or many interaction rules in play, we allowed the planner for EMPA (and all ablations described below) to run for as long as needed in real time to find a solution, subject to the node search budget for each move described above and a total time budget of 24 hours for playing each game.

\subsection*{Planning: intrinsic rewards}\label{intrinsic_rewards}

The agent's intrinsic reward function is a sum of the goal reward, subgoal reward and goal gradient reward: $V(s, \theta) = R_{G}(s, \theta) + R_{SG}(s, \theta) + R_{GG}(s,\theta)$.

\textbf{Goal Reward:} All planner modes return a plan if a goal state is reached, and they stop searching states that stem from loss states, so $R_{g}(s,\theta)$ is effectively $\infty$ for $\texttt{Win}$ states, $-\infty$ for $\texttt{Loss}$ states, and $0$ otherwise.

\textbf{Subgoal Reward:} By contrast, subgoal rewards only cause the short-term planner to return a plan, but they drive all planners' intrinsic rewards. Taking advantage of the fact that all termination goals specify conditions of the form $|c_{j}|=N$, states are penalized proportional to their distance from object-count goals. Specifically, the subgoal reward for a state $s$ is 
\begin{align}
R_{SG}(s, \theta) = \rho \sum_{g\in \theta_{G}} \frac{N_{g_{c}} - |c(g)|}{|c(g)|^{2}}(-1)^{\mathbbm{1}[g=Win]}
\end{align}
where $N_{g_{c}}$ is the class count specified by $g$ and $|c(g)|$ is the actual class count in the state, of the class specified in $g$. 
$\rho$ calibrates the relative value of subgoal reward to goal gradient reward (explained below); we use $\rho = 100$.

\textbf{Goal gradient reward:} The numeric subgoal reward operates at the abstract level of object counts. Changes in this quantity are too sparse to guide search on their own, so the planner uses ``goal gradients'' to score states at a lower level of abstraction by maximizing 
\begin{align}
R_{GG}(s,\theta) = \sum_{g\in \theta_{G}} \frac{d_{min}\big(c'(g), c(g)\big)}{|c(g)|^{2}}(-1)^{\mathbbm{1}[g=Win]}
\end{align}
where $c'(g)$ refers to the class (if one exists) that can destroy items of class $c(g)$, and $d_{min}(c_{j},c_{k}$) refers to the distance between the most proximal instances of classes $c_{j}$ and $c_{k}$. This biases the agent to seek proximity between objects that need to be destroyed and the objects that can destroy those objects, and vice-versa for objects that need to be preserved.

The specifics of the goal gradients (that is, which particular objects to approach, avoid, ignore, and so on) are generated by the planner within each game by analyzing EMPA's highest probability model. For example, if the highest probability model has ``The agent wins if there are 0 diamonds on the screen" as a (paraphrased) termination condition, the planner finds all classes that, according to the model, can remove diamonds from the screen, and gives high value to states in which such objects are near diamonds. More generally, if the highest probability model has $|c_{j}|=0$ as a \texttt{Win} condition for some class, $c_{j}$, the planner finds all classes $c_{k}$ that the model believes can destroy $c_{j}$ and generates goal gradients as specified above. This simply involves finding all the $c_k$ in rules $(c_j, c_k, \xi)$ in which the affected class is $c_{j}$ and the predicate, $\xi$, is one of \texttt{\{destroy, pickUp, transformTo\}}.

In addition to these goal-gradient heuristics, the planner adopts several special ways of treating resources and projectiles that are not strictly necessary, but are helpful given the use of a restricted planning budget (see Supplementary Information).

\subsection*{Planning: state pruning}
The planner ameliorates the exponential explosion of the state space by using Iterative Width (IW) \cite{geffner2012width,lipovetzky2017best}, which prunes states that are not sufficiently different from states previously encountered in the course of search. This is achieved by defining \emph{atoms}, logical propositions that can evaluate to \texttt{True} or \texttt{False}. We indicate the presence of each object in the state with the atom, (\texttt{object}, \texttt{True}), and we use (\texttt{object}, \texttt{False}) for a known object that is no longer in the game state. We encode the location of each object with the atom, (\texttt{object}, \texttt{posx}, \texttt{posy}). For the agent avatar, whose transition probabilities are additionally affected by its orientation, we use a (\texttt{object}, \texttt{posx}, \texttt{posy}, \texttt{orientation}) tuple. The algorithm maintains a table of all atoms that have been made true at some point in the search originating from a particular actual game state. During search, IW prunes any state that fails to make some atom true for the first time in the search.

In games with many moving objects, IW can experience sufficient novelty over a great many states without the agent needing to make a move; this leads to very inefficient search. To incentivize the agent to move, we additionally use a penalty on repeated agent positions, $\alpha \cdot$\texttt{count}(\texttt{agent}, \texttt{posx}, \texttt{posy}, \texttt{orientation)}$^{2}$, with $\alpha =-10$ when there are moving objects other than the avatar in the game; otherwise $\alpha =-1$. To further reduce IW's sensitivity to moving objects, we do not generate atoms for the locations of objects hypothesized to be random or whose presence in the game is predicted to last fewer than $5$ seconds, and do not track the location of projectiles produced by the agent.

\subsection*{Planning: prediction-error monitoring and re-planning}\label{replanning}
In all planning modes, the planner takes as input an $(s_{t}, \theta)$ pair and returns a list of high-value actions, $a_{t:N}$, together with their corresponding predicted states from the simulator, $\hat{s}_{t+1:N+1}$. When the agent takes an action $a_{t}$, the true environment returns $s_{t+1} \sim p_{a_{t}}(s_{t+1}, s_{t})$. After each action, the agent compares the true $s_{t+1}$ to the predicted $\hat{s}_{t+1}$. If the agent's own position is not as predicted, or if the agent is too close to a dangerous object whose position was not as predicted, the planner is run again; otherwise, the agent continues execution of the action sequence. We defined ``close'' as within 3 game squares, but any distance greater than 3 would work as well, at the cost of more frequent re-planning.

Because planning and acting without a confident dynamics model will almost always lead to immediate prediction errors and re-planning, the agent waits for a small number of time-steps before attempting to move in a new game, and at the beginning of a new level, while it observes the motions of any dynamic objects in that level. We set these waiting thresholds at 15 and 5 steps, respectively, but any similarly low values would work as well.

\subsection*{DDQN Implementation}
We ran DDQN (based on the public repository \url{https://github.com/dxyang/DQN_pytorch}) with parameter settings %
$\alpha=0.00025$, $\gamma=0.999$, $\tau=100$, experience-replay max $=50,000$, batch size $=32$, and image input recrop size $=64\times64\times3$. For exploration, $\epsilon$ was annealed linearly from 1 to .1 over either $1,000$, $10,000$, or $100,000$ steps. %
Each model was run for a maximum of $1$ million agent steps. Annealing over $100,000$ steps performed best, and these results (averaged across three independent runs, each with a different random seed) are reported in the main text. 

\subsection*{Rainbow Implementation}
We ran Rainbow (based on the repository \url{https://github.com/google/dopamine} \cite{castro18dopamine}) with parameter settings optimized for sample-efficient performance taken from \cite{kaiser2019model}: %
num-atoms $=51$, vmax $=10$, $\gamma=0.99$, update-horizon $=3$, min-replay-history $=20,000$, target-update-period $=50$, update-period $=1$, epsilon-train $=0.01$, replay-scheme $=$ ``prioritized'', optimizer-learning-rate $=0.0000625$, optimizer-epsilon $=0.00015$.
We found that an annealing period for $\epsilon$ of $150,000$ steps produced greater sample efficiency, and report these results (averaged across three independent runs, each with a different random seed) in the main text. Each model was run for a maximum of $1$ million agent steps.

\subsection*{EMPA ablations: exploration}

We considered two $\epsilon$-greedy variants to EMPA's exploration. In both variants, the planner ignores exploration goals and only pursues win-related goals. The first variant executes a random action with probability $\epsilon$, and executes the action suggested by the planner with probability $1-\epsilon$. The second $\epsilon$-greedy DF variant executes the same random-exploration policy, but re-sets the long-term planning node budget to its initial value after any level forfeit. We annealed $\epsilon$ linearly from $1$ to $0.1$ for both variants, and ran each variant with both a $1000$ and $2000$ step annealing schedule.

\subsection*{EMPA ablations: planning}
We evaluated five versions of EMPA with ablations to the planner. Three of these ablations removed representations of subgoals, goal gradients, or both subgoals and goal gradients, which affect both the intrinsic reward function as well as the planner's completion criteria. We also considered an ablation without IW state pruning, and finally an ablation that lacks all three of these features.

\subsection*{Learning efficiency}

Confidence intervals shown in Figure \ref{fig:composite_ratio} were generated by calculating the efficiency ratio of the mean EMPA run to the mean human participant over 10,000 bootstrapped samples for each game, with the exception of two games (Boulderdash 2 and Sokoban) on which EMPA obtained an efficiency score greater than 0 fewer than three times; in those cases only human participants were resampled.

\subsection*{Behavioral analyses}

Fine-grained behavioral analyses in Figure \ref{fig:behavioral_analysis} were generated by classifying objects from 78 games into the following categories: 
Positive: if instances of this class participate in a win condition and the avatar can destroy these objects, or if touching instances of this class grants points.
Instrumental: if touching an instance of this class is necessary for a solution or leads to a significantly shorter solution than is possible without touching the item.
Negative: if touching an instance of this class results in a loss, loss of points, or significantly delays a solution.
Neutral: if none of the above obtain.

\clearpage

\section*{Supplementary Information}

\subsection*{Game Descriptions}

\hspace{\parindent}
\textbf{Antagonist} 
The player has to eat all the hot dogs before the antagonists eat the burger. The player can win by playing keepaway with the burger while trying to reach the hot dogs, but in the last level there are too many hot dogs to eat and only one burger --- the only way to win is to push the burger onto a piece of land that the antagonists cannot get to.

\textbf{Antagonist 1}
The player can no longer play keepaway with the burger, but can push boxes into the antagonists' path to block them.

\textbf{Antagonist 2}
The hot dogs are harder to get to. To make things a bit easier, the player can put frosting on the burgers, which makes the antagonists not recognize their food. There's a smarter antagonist that isn't hungry but takes the frosting off the burgers, making them recognizable again. The player has to balance frosting burgers and eating hotdogs in order to win.

\textbf{Aliens}
(GVGAI) This is the VGDL version of the classic ``Space Invaders''. The player moves a spaceship left and right and can shoot surface-to-air missiles. Alien bombers move side-to-side and drop bombs at the agent, who can hide under a base. The base slowly gets destroyed, so the player must often dodge bombs while shooting all the aliens.

\textbf{Aliens --- Variant 1} The bombs shot by the aliens move randomly.

\textbf{Aliens --- Variant 2}
The way to win now is to destroy the protective base.

\textbf{Aliens --- Variant 3} The player's surface-to-air missiles move three times as slowly, making aiming more challenging.

\textbf{Aliens --- Variant 4} Aliens move three times as fast, shoot bombs six times as often, and the player's base no longer destroys incoming bombs. To compensate, the player can now shoot surface-to-air missiles continuously.

\textbf{Avoidgeorge}
(GVGAI) Evil George chases citizens. If he touches them, they become annoyed, and if there are no calm citizens in the game, the player loses. To avoid this, the player can feed the annoyed citizens candy, which makes them calm down. If George touches the player, the player dies. Keeping citizens calm for 500 game steps results in a win. 

\textbf{Avoidgeorge --- Variant 1}
The player can now use the candy to make annoyed citizens disappear, and if all annoyed citizens disappear, the player wins.

\textbf{Avoidgeorge --- Variant 2}
The player can still make annoyed citizens disappear, and if all annoyed citizens disappear, the player wins. George moves faster, but the player can now throw the candy, making it easier to reach the citizens.

\textbf{Avoidgeorge --- Variant 3}
Same as Variant 2, but the player begins the game stuck in a room and needs to tunnel out of it before doing anything else.

\textbf{Avoidgeorge --- Variant 4}
Same as Variant 3, but George moves even faster, and the candy no longer makes citizens disappear.

\textbf{Bait}
(GVGAI) A puzzle game. The player has to pick up a key to go through a door to win. Falling into holes in the ground results in death, but pushing a box into a hole covers it, destroys the box, and clears the path.

\textbf{Bait --- Variant 1}
Same as the original, but there are dirt cannons lying around that fill any holes in the dirt's path. Filling the wrong holes can make it impossible to remove certain boxes from the game board.

\textbf{Bait --- Variant 2}
In some levels, the player has to make the key by pushing metal into a mold. In other levels, the normal key exists, but is harder to reach than are the metal and the mold.

\textbf{Bees and Birds} The player has to get the goal while avoiding swarming bees. In level 2, the goal is surrounded by an electric fence, but the player can release a bear that can eat through the fence. In level 3, the player can either go through a dangerous swarm of bees or take a longer path that involves releasing a bear that eats the fence. In level 4, the player can get to the goal simply by going around the fence. Releasing the bear will also cause the bees to be released, making winning more challenging.

\textbf{Bees and Birds --- Variant 1} The player has to get the goal while avoiding swarming bees. In all subsequent levels, there are lots of swarming bees, so the best stategy is to release a mockingbird that will eat the bees, clearing the path to the goal.

\textbf{Boulderdash}
(GVGAI) This is the VGDL version of the classic ``Dig Dug''. The player has to pick up 9 diamonds and then exit the level. Boulders in the game are supported by dirt, and if a boulder falls on the player, the player dies. The player has to avoid two types of subterranean enemies to avoid dying.

\textbf{Boulderdash --- Variant 1} Same as the original, except the player only has to pick up three diamonds to win.

\textbf{Boulderdash --- Variant 2} Same as the original, except that one of the subterranean enemies converts dirt to diamonds instead of being trapped by it.

\textbf{Butterflies} The player has to catch all the butterflies before the butterflies activate all the cocoons. The butterflies move randomly, and if they touch a cocoon, the cocoon turns into a butterfly.

\textbf{Butterflies --- Variant 1} Same as the original, but butterflies are faster.

\textbf{Butterflies --- Variant 2} There are more cocoons, meaning there will soon be many more butterflies. The player can destroy cocoons, but still dies if all cocoons are cleared.

\textbf{Chase}
(GVGAI) The player has to chase all the birds, which flee the player. Touching a bird turns it into a carcass. If a scared bird touches a carcass, the carcass turns into a predator bird which chases and can kill the player.

\textbf{Chase --- Variant 1}
Same as the original, except that there are a few gates that release predator birds sporadically.

\textbf{Chase --- Variant 2}
Touching a bird kills it, rather than turning it into a carcass. The game now contains a gate, which if the player shoots, releases a wolf that quickly chases and kills the player.

\textbf{Chase --- Variant 3} In addition to the birds, there are sheep that rapidly pace in a predetermined path. If these touch a carcass, they turn into zombies that chase the player.

\textbf{Closing Gates} The player has to get to the exit before large sliding gates close.

\textbf{Closing Gates --- Variant 1} The gates close more quickly. The only way for the player to escape is to shoot a bullet onto the gates' path; this blocks the gates and lets the player squeeze through.

\textbf{Corridor}
The player has to get to the exit at the end of a long corridor while avoiding fireballs of different speeds that fly toward the player.

\textbf{Corridor --- Variant 1}
The player can now shoot bullets at the fireballs. Depending on the speed of a fireball, the bullet will either slow it down or speed it up (in fact, the speed of any fireball can be toggled in a cycle by hitting it with more bullets).

\textbf{Explore/Exploit} 
The board is full of colorful gems, and the player wins by picking up all gems of any given color. Different board configurations incentivize exploring new ways of winning versus exploiting known ways.

\textbf{Explore/Exploit --- Variant 1} Same as the original game, except evil gnomes chase the player as the player collects the gems.

\textbf{Explore/Exploit --- Variant 2}
There are no evil gnomes, but now the gems are being carried by gnomes that flee the player.

\textbf{Explore/Exploit --- Variant 3}
Same as the original, except one of the gems the player encounters early on is poisonous.

\textbf{Helper} The player has to help minions get to their food. Sometimes the food is blocked by a boiling pot of water that only the player can destroy, and sometimes the minions are boxed in by a fence that the player can push food through.

\textbf{Helper --- Variant 1}
The player wins by eating all the food, but can only eat it after taking it to the minions for processing.

\textbf{Helper --- Variant 2}
The player's goal is once again to feed minions. In this variant the player can shoot a path through red fences to free minions. In the last level, the player has manually clear one path, then shoot through some fences, and then push food through a third fence in order to feed the minion.

\textbf{Frogs}
(GVGAI) This is the classic ``Frogger''. The player has to cross a road while avoiding dangerous cars and then step carefully on moving logs to cross a river, to get to an exit. The level layouts of this game are slightly modified from the original GVGAI layouts, in order to make the game playable in the original version of VGDL this project was built on.

\textbf{Frogs --- Variant 1}
The cars move differently now; they move at different speeds, and when they hit the edge of the screen they rapidly turn around and drive in the other direction.

\textbf{Frogs --- Variant 2}
The cars and logs move faster than the original version, and fewer logs float down the river, making it more difficult to reach the goal by ordinary means. But there is a device that, if used, teleports the player nearer to the goal.

\textbf{Frogs --- Variant 3}
Now there aren't any logs to help the player cross the water, but the player can throw mud at the water to build bridges.

\textbf{Jaws}
(GVGAI) The player dies if touched by a chasing shark. Cannons on the side of the screen shoot cannonballs at the player. The player can shoot bullets at the cannonballs, to convert them into a new kind of metal. Picking this metal up gives points. The player wins by surviving for 500 game steps.
 
\textbf{Jaws --- Variant 1}
Now there is a fence that prevents the player from getting too close to the cannons.

\textbf{Jaws --- Variant 2}
Each piece of metal grants 5 health points; having any health points protects the player from the shark; the shark takes away one health point each time there's contact. The shark can be destroyed if the player has 15 health points. The player wins after destroying the shark or surviving for 500 game steps.

\textbf{Lemmings} 
(GVGAI) In this classic game, the player has to help a group of lemmings get to their own exit. To do so, the player must shovel a tunnel through dirt, and incur a loss of points with every shovel action.

\textbf{Lemmings --- Variant 1}
Now the player gets points for shoveling, rather than losing points.

\textbf{Lemmings --- Variant 2}
The player can release a mole, which loves to shovel and will clear all the dirt. But if the mole happens to touch the player, the player dies.

\textbf{Lemmings --- Variant 3}
The mole is still there and will shovel dirt when released. But if the mole touches a lemming, the mole turns into a snake that chases and kills the player.

\textbf{Missile Command} (GVGAI) Spaceships want to destroy the player's bases. The player defends the bases by shooting short-range lasers, and wins upon destroying all the spaceships, or loses after all the bases are destroyed.

\textbf{Missile Command --- Variant 1} The bases don't stay in place; they float around randomly.

\textbf{Missile Command --- Variant 2} The player can now shoot long-range lasers. However, these lasers bounce off the edges of the game and kill the player upon contact.

\textbf{Missile Command --- Variant 3} 
A newer, faster spaceship tries to destroy the player's base, and the player is only equipped with a short-range laser.

\textbf{Missile Command --- Variant 4}
There are more fast spaceships, making it very difficult to destroy all of them before they reach the bases. However, the player can shoot the short-range laser at the ozone layer, which transforms into a shield that the enemies cannot pass through.

\textbf{MyAliens}
(GVGAI) The player can only move sideways. Fast-moving bombs drop from the top. One type of bomb kills the player; the other gives points. Surviving the onslaught of bombs for 500 game steps results in a win.

\textbf{MyAliens --- Variant 1}
The player can now collect the safe bombs, and can go to an exit once five of those have been collected.

\textbf{MyAliens --- Variant 1}
The player can now move in all directions, but bombs come from all directions, too. The player wins by either picking up five safe bombs and getting to the exit, or by surviving for 500 game steps.

\textbf{Plaqueattack}
(GVGAI) 40 Burgers and hotdogs emerge from their bases and attack cavities; if they reach all the cavities the player dies. The player wins by destroying all the hotdogs and burgers with a laser.

\textbf{Plaqueattack --- Variant 1} The burgers and hotdogs move faster, but there are only 24 of them. The player has to destroy them while avoiding a roving drill.

\textbf{Plaqueattack --- Variant 2} There is no drill, but now there are bouncing projectiles that kill the player on contact.

\textbf{Plaqueattack --- Variant 3} Now the player has to face 40 burgers and hotdogs again, as well as the bouncing projectiles, but can win by destroying all the gold fillings, instead.

\textbf{Portals}
(GVGAI) The player wins by reaching an exit. Certain rooms, including the room with the exit, are accessible only by going through portals that teleport the player to their portal-exits. Most portals have multiple portal-exits, and the player's final location is randomly chosen from those. While completing this task, the player has to avoid bouncing missiles, roving space-monsters, and inert traps.

\textbf{Portals --- Variant 1}
Same dynamics as above, with a new portal type and more complicated level layouts, but with fewer missiles and space-monsters.

\textbf{Portals --- Variant 2}
Uses the complex layouts as in Variant 1. The player now has to contend with roving monsters that can pass through walls.

\textbf{Preconditions} 
The player wins by picking up a diamond. Getting the differently-colored fake diamond does nothing. Usually the diamond is surrounded by poison, which the player can pass through only after drinking an antidote. In the last level the player has to drink an antidote to pass through the poison that blocks two antidotes, in order to pass through the double-poison in order to get to the triple antidote, in order to get through the triple poison that guards the diamond.

\textbf{Preconditions --- Variant 1}
The player can also pick up a more powerful antidote that allows passage through two poisons. Different level layouts encourage using either the single or double antidote.

\textbf{Preconditions --- Variant 2}
Same rules as the original, except only one antidote can be ingested at a time.

\textbf{Push Boulders} The goal is to get to the exit. Touching silver or green ore results in a loss. Boulders can be pushed by the player; these destroy the ore. Limestone can be destroyed by the player but otherwise has no effect on the game. The game often involves solving maze-like challenges and necessitates using boulders to clear dangerous obstacles.

\textbf{Push Boulders --- Variant 1} Boulders destroy green (but they don't destroy or push silver). The mazes are more difficult and involve multiple uses of boulders to clear ore.

\textbf{Push Boulders --- Variant 2} Now the boulders are too heavy to move, but one-time-use-only cannons shoot cannon balls that destroy any boulders or ore in their path. The player has to fire up the right cannons to clear paths. Dust clouds do nothing.

\textbf{Relational} The goal is to make all the blue potions disappear, which happens when they are pushed into fire. In the second level, there is no fire, but touching the red potion converts it to fire. In the third level there is no red potion or fire, but pushing a box into a purple potion converts the two into fire. In the fourth level, a green potion can be converted into a box that can be pushed into a purple potion to make the fire.

\textbf{Relational --- Variant 1} Same relational rules as above, but the blue potion is now carried by a gnome that chases the player. While nothing happens if the gnome touches the player, the player has to move around in such a way as to get the gnome to run into fire.

\textbf{Relational --- Variant 2} Now the gnome is fleeing the player. While nothing happens if the player touches the gnome, the player has to move around in such a way as to get the gnome to run into fire.

\textbf{Sokoban}
(GVGAI) In this classic puzzle game, the player wins by pushing boxes into holes.

\textbf{Sokoban --- Variant 1}
The player can make use of portals, which teleport either the player or boxes to a particular portal-exit location on the game screen.

\textbf{Sokoban --- Variant 2}
This variant does not have portals. Instead, there is dirt lying around that can fill holes. Filling certain holes makes the levels unwinnable.

\textbf{Surprise}
The player has to pick up all of the red apples. Touching a cage releases a randomly-moving Tasmanian devil. If the Tasmanian devil touches a green gem, the Tasmanian devil gets cloned and the gem disappears. The player can pass through gems and Tasmanian devils

\textbf{Surprise --- Variant 1}
The player can no longer pass through the Tasmanian devils; releasing the devils makes winning much more challenging.

\textbf{Surprise --- Variant 2}
The goal is to destroy all the gems, which can only be done by releasing the Tasmanian devil.

\textbf{Survive Zombies}
(GVGAI) The player has to survive for 500 game steps while avoiding zombies that emerge from fiery gates. The fiery gates are dangerous, too. Fortunately, bees come out of flowers, and when bees touch zombies, the two collide and turn into honey. Eating honey gives the player temporary immunity from zombies.

\textbf{Survive Zombies --- Variant 1} The way to win is to collect all the honey.

\textbf{Survive Zombies --- Variant 2} The way to win is now to kill all the zombies, by first eating some honey for immunity.

\textbf{Watergame}
(GVGAI) A puzzle game. The player has to reach an exit, which is usually surrounded by water that drowns the player. Pushing dirt onto water clears the water.

\textbf{Watergame --- Variant 1} Red boulders can be pushed onto the dirt, thereby destroying it. Destroying some dirt is necessary in order to clear space, but destroying the wrong pieces of dirt makes levels unwinnable.

\textbf{Watergame --- Variant 2} More dirt can be made by mixing light-green and yellow potions.

\textbf{Zelda}
(GVGAI) The player has to pick up a key and get to a door, while using a sword to destroy randomly-moving dangerous creatures.

\textbf{Zelda --- Variant 1}
The player has to pick up three keys before getting to the door.

\textbf{Zelda --- Variant 2}
The player only needs one key, now, but both the key and the door move randomly throughout the level.

\textbf{Zelda --- Variant 3}
Now the key and door are carried by elves that flee the player.

\clearpage

\subsection*{Resource and Projectile Heuristics}

\begin{itemize}
  \item {\bf Picking up resources is a subgoal:} Conditional interactions in our version of VGDL occur as a function of either having 1 instance of a resource or having as many as the agent can carry. Both of these are treated as planner subgoals.
\end{itemize}

\begin{itemize}
  \item {\bf Transfer goal gradients from “Flicker’’ objects emitted by the agent to the agent itself:} Such objects (e.g., swords) appear directly in front of the avatar and disappear shortly after use. Since the avatar must be near an object in order to make contact between the Flicker and that object, this heuristic incentivizes the avatar to approach objects it wants to contact with the Flicker.
  \item {\bf Use rollouts when the agent fires a projectile:} The intrinsic reward for each action is ordinarily calculated as a function of the immediately-resulting state. When the agent shoots what it believes to be a projectile, it instead simulates random actions forward until that projectile is no longer on the screen or until N steps have passed (where N is equivalent to the block length of the longest axis on the game screen). Any subgoals reached during the rollout that involve the projectile are considered to be subgoals achieved by the original action of shooting.
  \item{\bf Do not explore the consequences of agent motion when exploring the consequences of firing projectiles:} In the normal course of search, when the agent explores the part of the search tree in which it has shot a projectile that is still on the game board, it only takes the no-op action as long as it is at least 3 block lengths away from the nearest object thought to be dangerous.
\end{itemize}
\clearpage

\begin{figure}[H]
\centering
\label{fig:vignette_two_games}
\vspace{-1.5cm}
\includegraphics[width=1\textwidth]{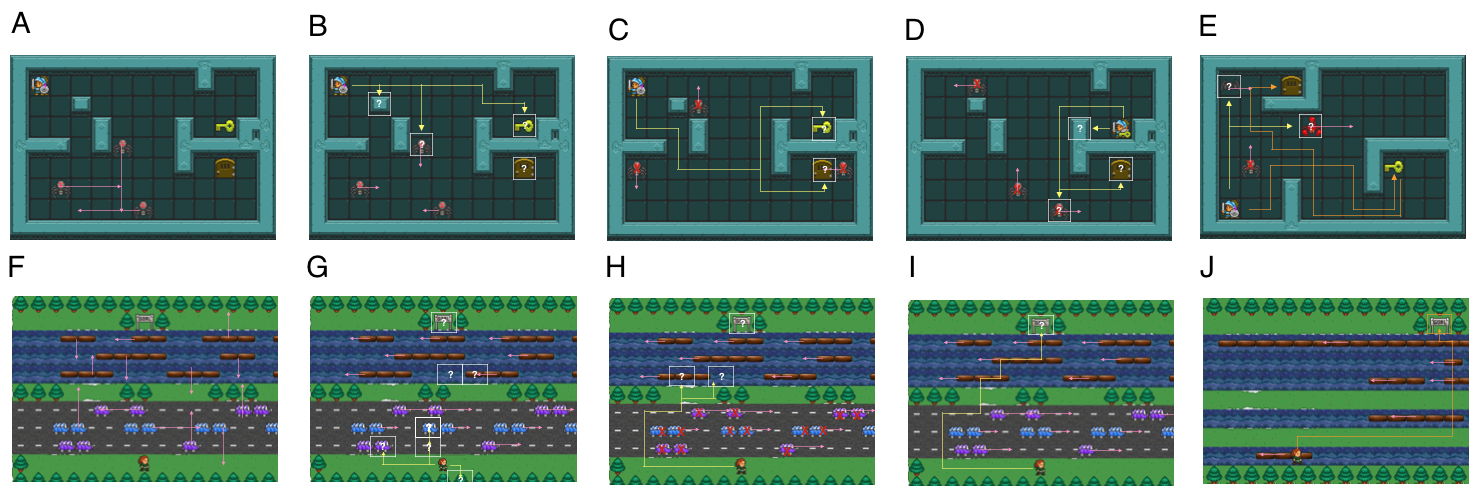}
\caption*{
{\bf Extended Data Figure 1:}
EMPA learning to play two games (Zelda and Frogs), showing human-like exploration and planning behaviors. Solving these (and many other) tasks requires first learning about objects that are difficult to reach using simpler stochastic exploration policies
\cite{mnih2015human, pathak2017curiosity,haber2018learning}: In order to achieve the first win in Zelda, the agent has to learn to pick up a key and then use it to open a door; in order to achieve the first win in Frogs, the agent has to learn to cross dangerous water by stepping on moving logs. EMPA, like humans, learns about these goal-relevant objects rapidly %
because it explicitly specifies reaching unknown objects as exploratory goals for the planner, rather than waiting to encounter them by chance as in $\epsilon$-greedy exploration \cite{mnih2015human} or surprise-based curiosity \cite{pathak2017curiosity,haber2018learning}. (A) In Zelda, EMPA begins by observing that spiders move but has not yet learned %
how they move, so it predicts random motions (pink arrows). %
(B) After several time-steps, EMPA learns to predict their motions %
and now begins to plan. It does not know %
what occurs upon contact with any of the objects, so it %
sets as exploration goals (white squares) to reach an object of each type. The planner finds efficient paths (yellow) to those objects and chooses whichever is found first. (C) After several dozen time-steps, the agent has learned that walls block motion and that spiders are dangerous (red x-shapes) but is still uncertain about the key and the door, so it plans paths to these exploration goals while avoiding the spiders. (D) After a few hundred time-steps, the agent has learned that it can pick up a key, which as a resource changes the state of the agent. This resets all exploration goals: maybe a key allows the agent to pass through walls, or makes it immune to spiders? (E) After another hundred steps, the agent has interacted with all objects both before and after touching the key, and has learned that the only effect of touching the key is that now touching the door wins the level. The agent's knowledge generalizes to any board layout that contains the same object types. Here it immediately plans a winning path (orange) to pick up the key and reach the door, though it may also generate paths to exploration goals (yellow) to new object types that appear, such as the bat and the scorpion.
(F-H) EMPA begins learning in Frogs just as in Zelda: it initially predicts random motion of moving objects until it receives enough evidence to predict the regular motion of the cars and logs. It then learns that cars and water are dangerous but that standing on a log gives it the ability to safely cross the water. (I) Equipped with this knowledge, EMPA can generate a path that allows it to fulfill its final exploration goal of reaching the flag. (J) Having learned to win the game, the agent now generates winning paths (orange) on entirely new game levels, although it may also generate paths to exploration goals (yellow) if new object types appear. 
}
\end{figure}

\begin{figure}[H]
\centering
\vspace{-2cm}
\includegraphics[width=1\textwidth]{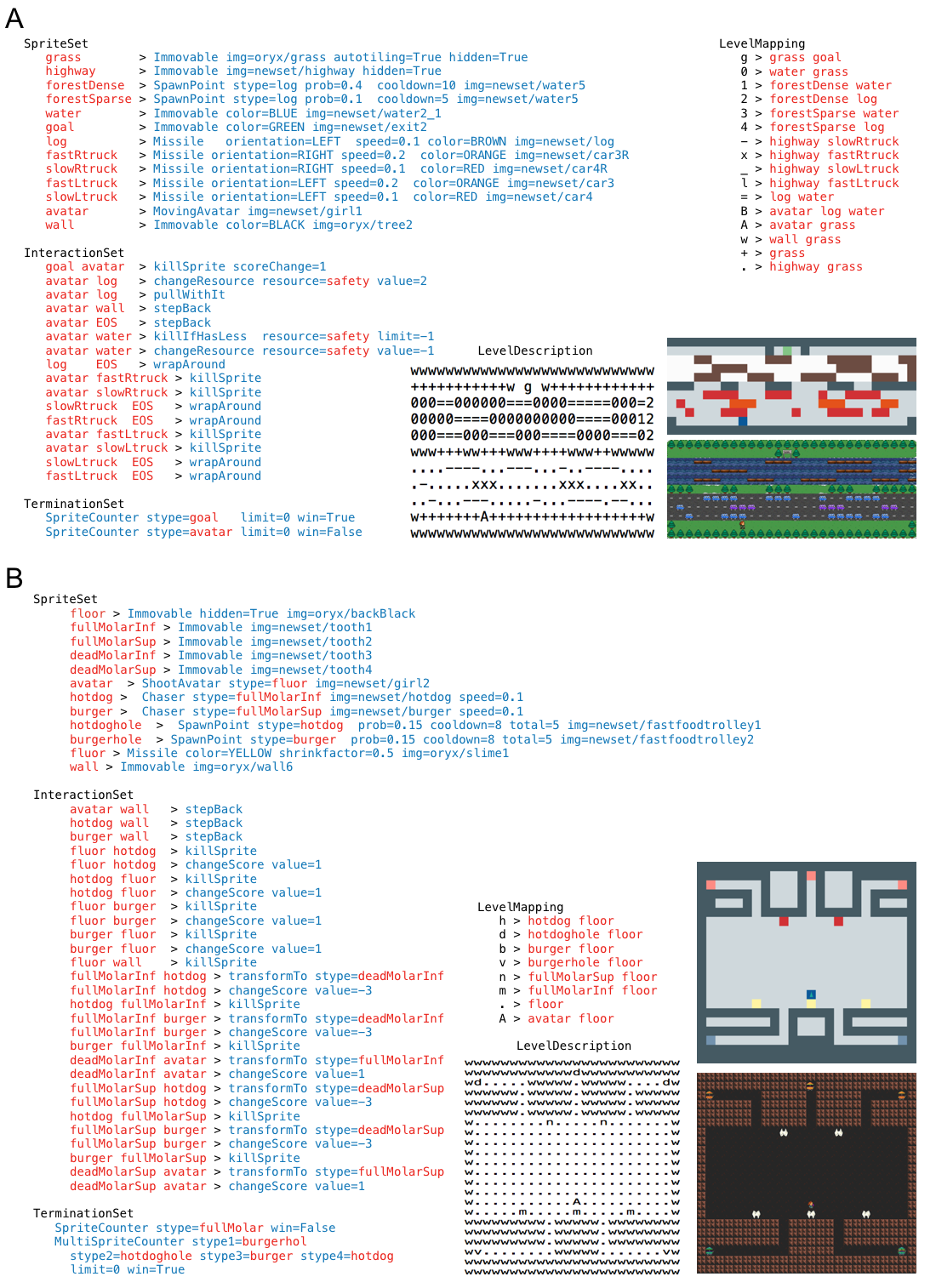}
\caption*{
{\bf Extended Data Figure 2:} VGDL descriptions of ``Frogs'' and ``Plaqueattack'', together with screenshots of representative levels in ``color-only'' and ``normal'' mode.
\label{fig:vgdl_description_frogs} 
}
\end{figure}
\clearpage

\begin{figure}[H]
\centering
\vspace{-2cm}
\includegraphics[width=1\textwidth]{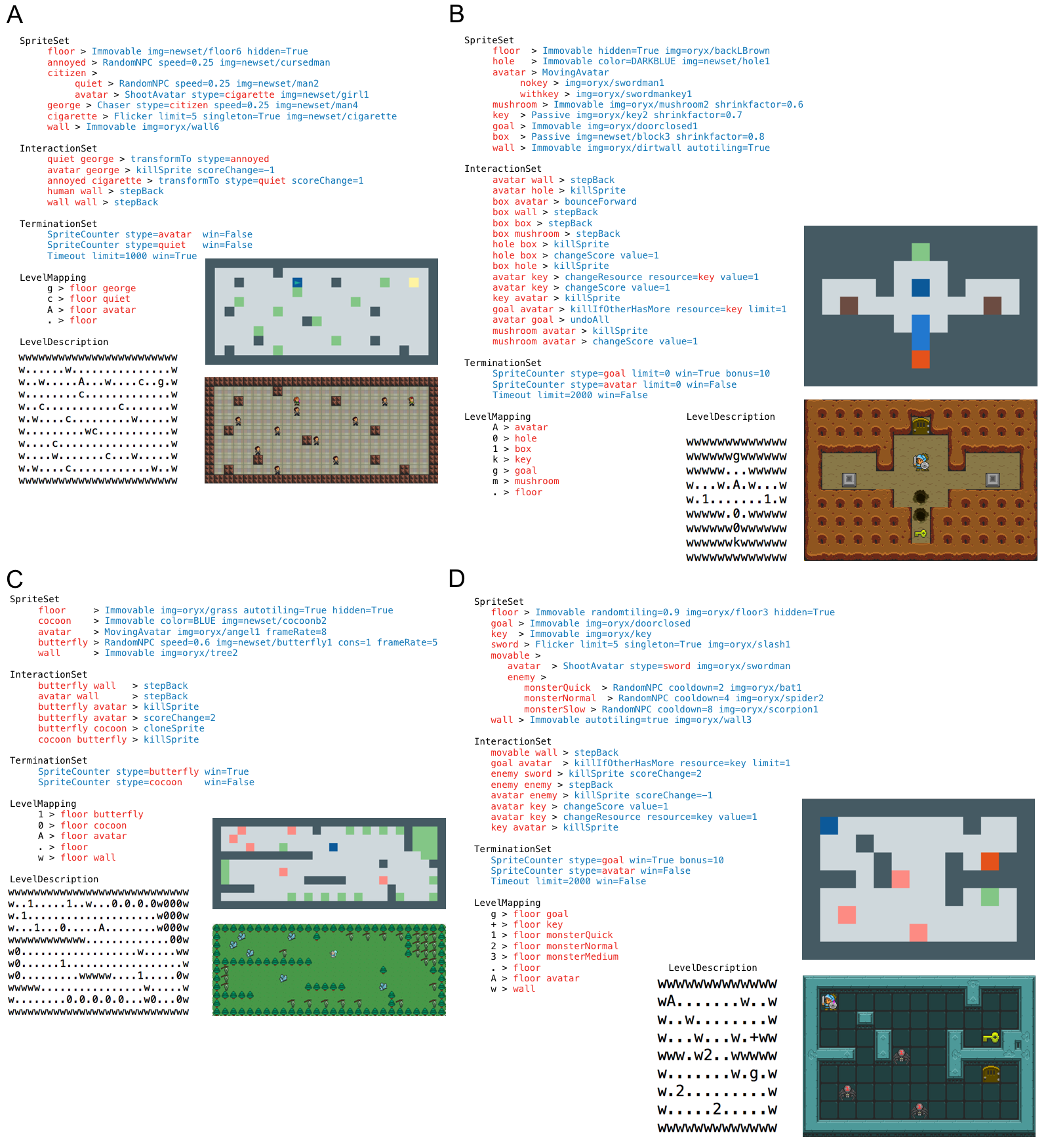}
\caption*{
{\bf Extended Data Figure 3:} VGDL descriptions of ``Avoidgeorge'', ``Bait'', ``Butterflies'', and ``Zelda'', together with screenshots of representative levels in ``color-only'' and ``normal'' mode.
\label{fig:vgdl_description_avoidgeorge} 
}
\end{figure}
\clearpage

\begin{figure}[H]
\centering
\includegraphics[width=1\textwidth]{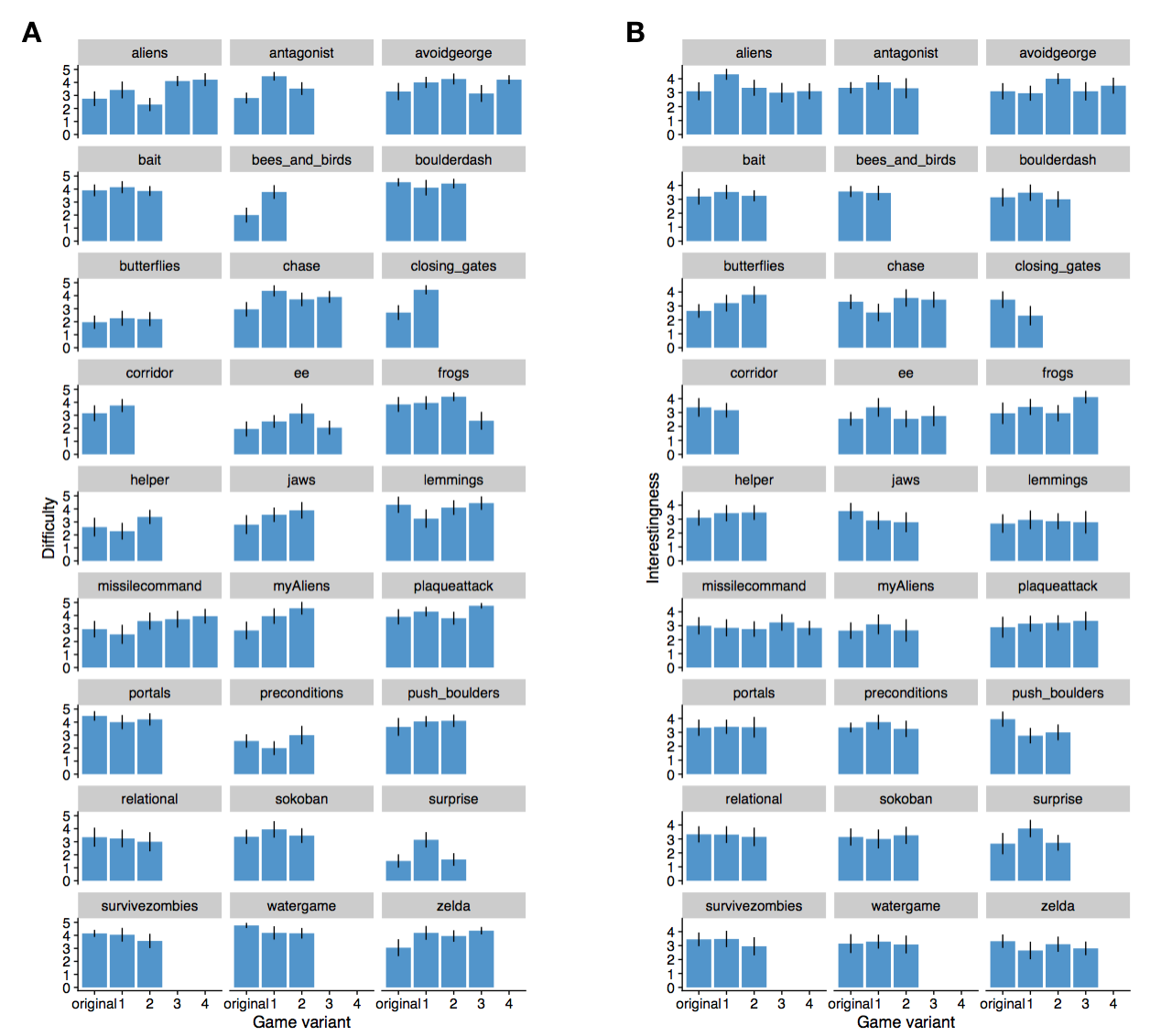}
\caption*{
\label{fig:game_ratings} 
{\bf Extended Data Figure 4:} Mean and $95\%$ confidence intervals of each game's subjective difficulty (panel A) and ``interestingness'' (panel B), grouped by source game. Our variant games are comparable to their source games on both metrics.
}
\end{figure}

\begin{figure}
\centering
\vspace*{-1cm}
\includegraphics[height=1\textheight]{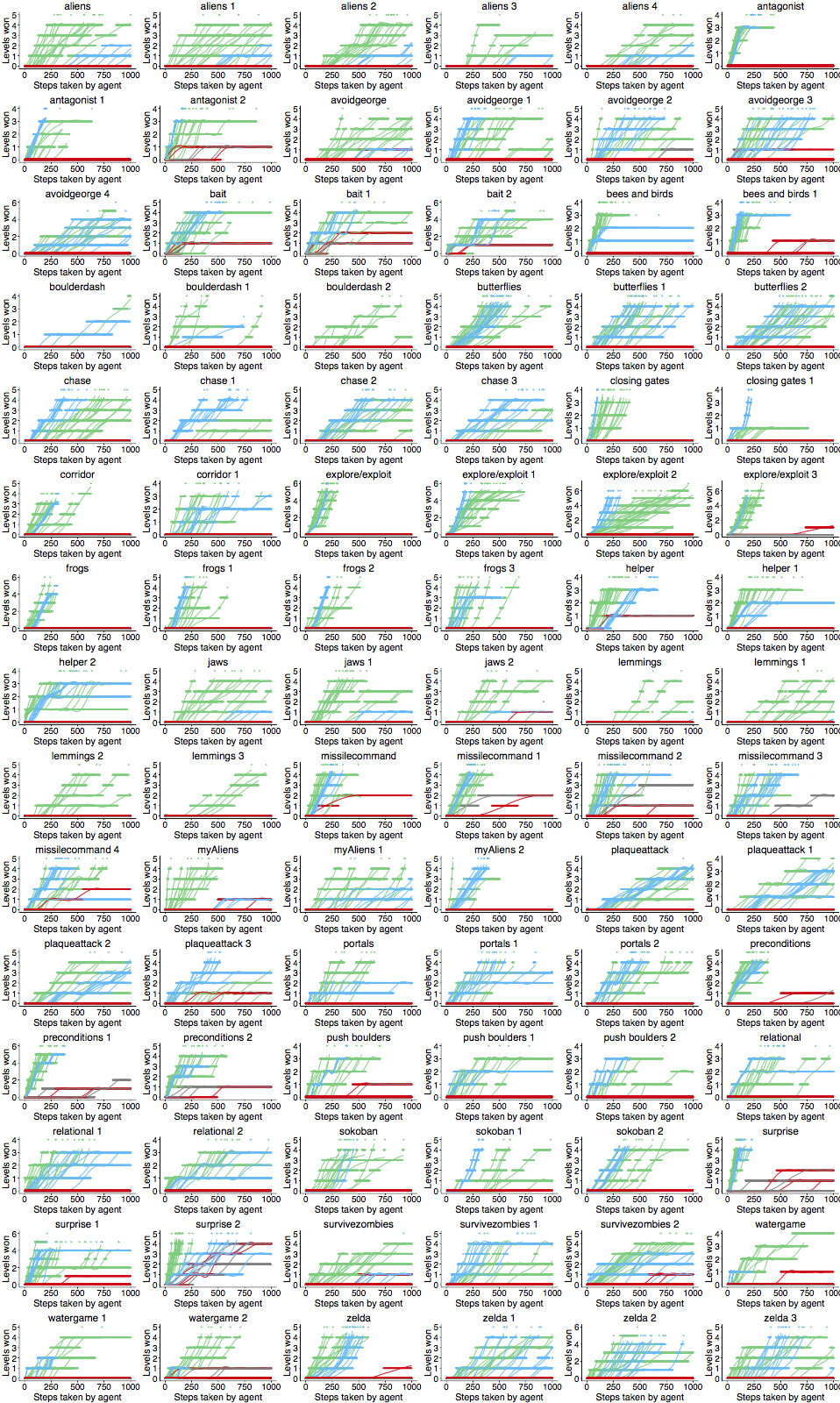}
\caption*{\label{fig:wins_1k}
{\bf Extended Data Figure 5:} Humans (green), EMPA (blue), DDQN (grey), and Rainbow (red) learning curves over the initial 1,000 steps of play.}
\end{figure}
\clearpage

\begin{figure}
\centering
\vspace*{-1cm}
\includegraphics[height=1\textheight]{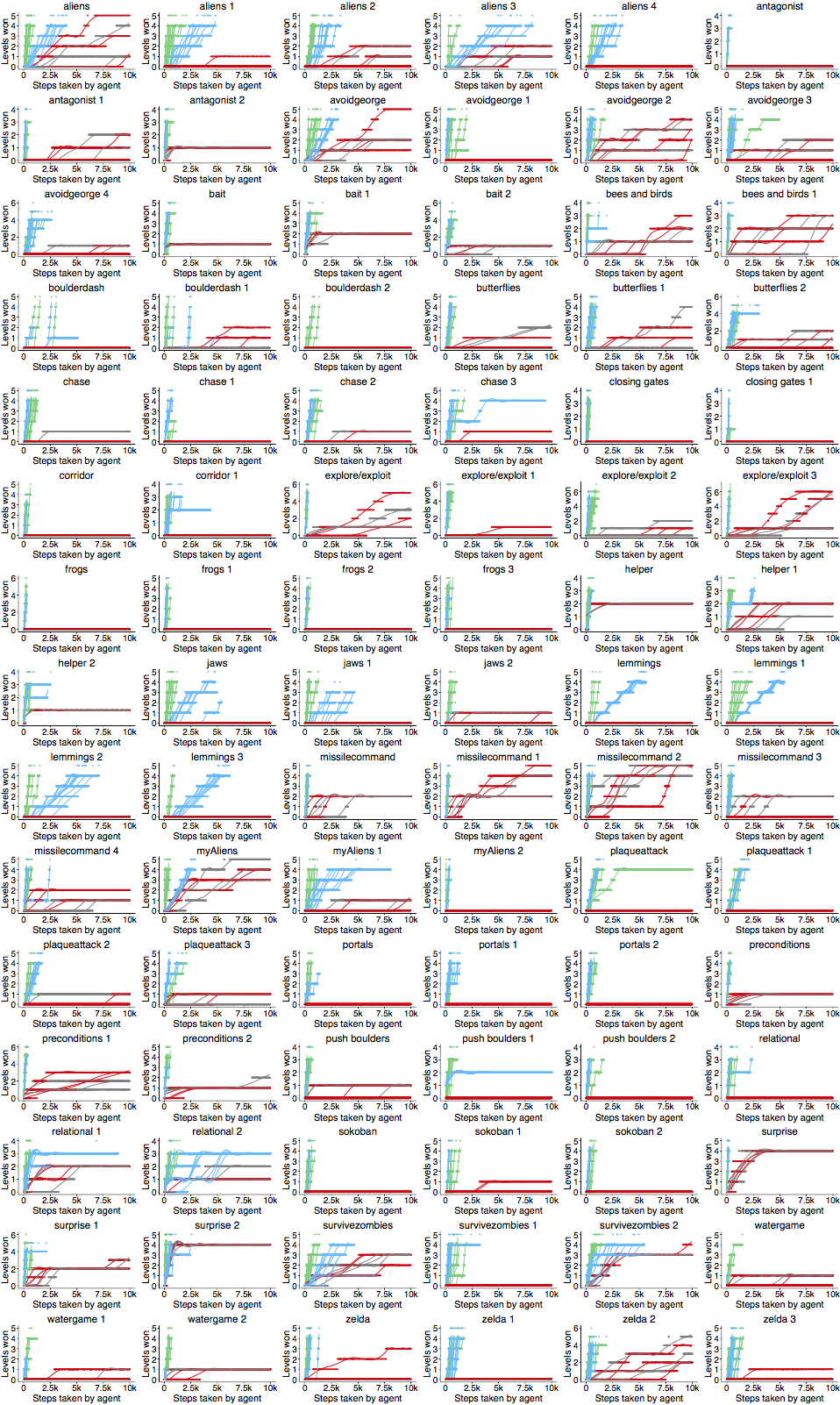}
\caption*{\label{fig:wins_10k}
{\bf Extended Data Figure 6:} Humans (green), EMPA (blue), DDQN (grey), and Rainbow (red) learning curves over the initial 10,000 steps of play.}
\end{figure}
\clearpage

\begin{figure}
\centering
\vspace*{-1cm}
\includegraphics[height=1\textheight]{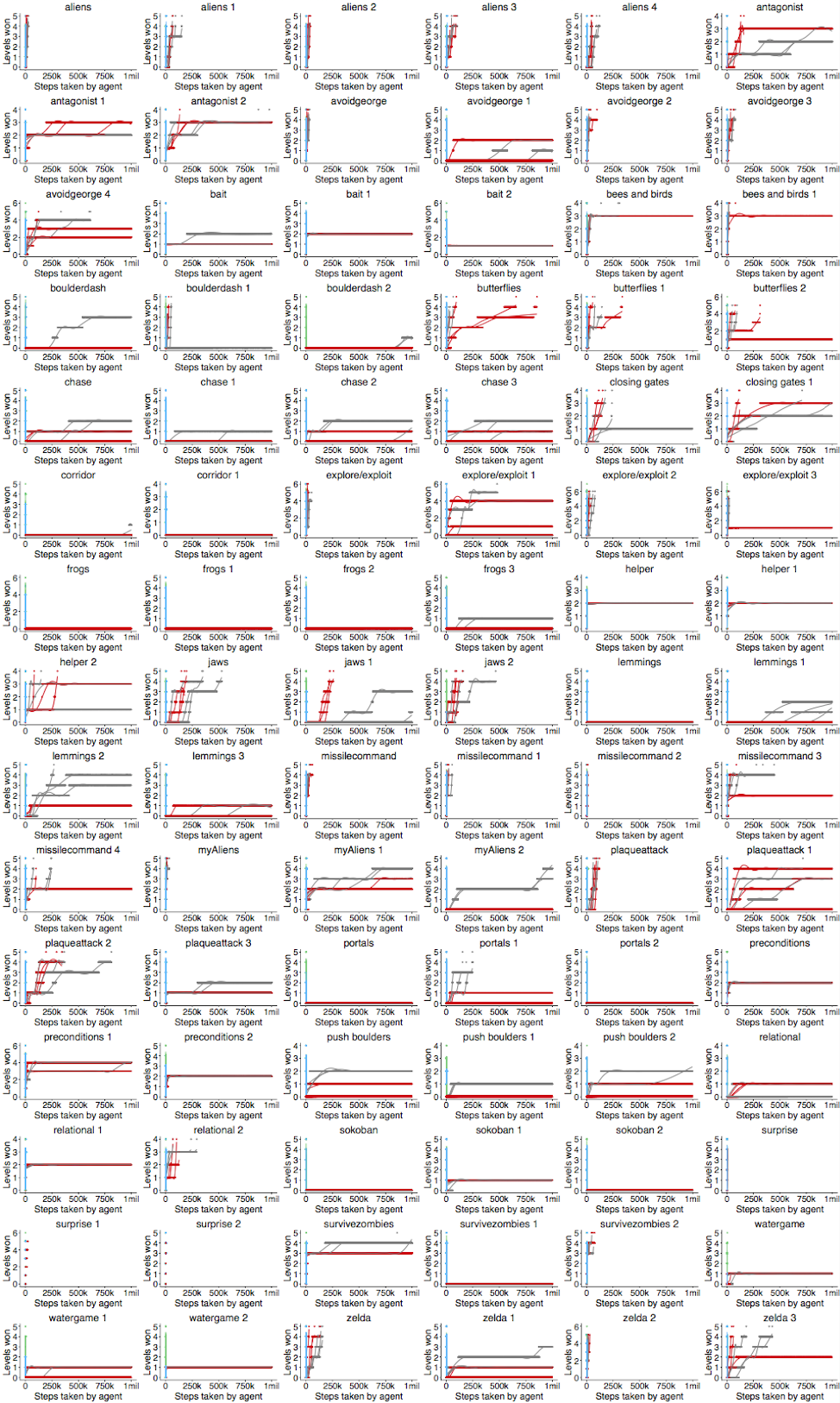}
\caption*{\label{fig:wins_1mil}
{\bf Extended Data Figure 7:} Humans (green), EMPA (blue), DDQN (grey), and Rainbow (red) learning curves over the initial $1$ million steps of play.}
\end{figure}
\clearpage

\begin{figure}[H]
\centering
\includegraphics[width=1\textwidth]{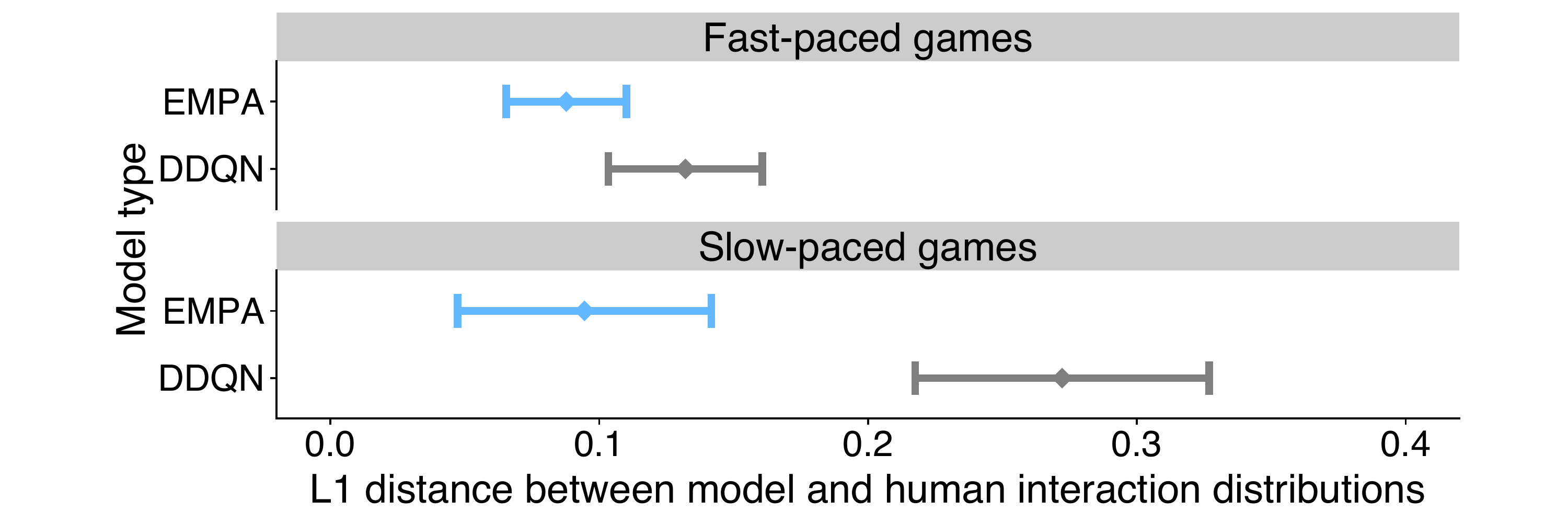}
\caption*{
{\bf Extended Data Figure 8:}
EMPA's distributions of object interactions match human distributions better than those of DDQN, particularly in slow-paced games in which the agent is the only moving item. Diamonds show mean distances; lines show 95\% confidence intervals.}
\end{figure}

\clearpage

\begin{figure}[H]
\centering
\includegraphics[width=1\textwidth]{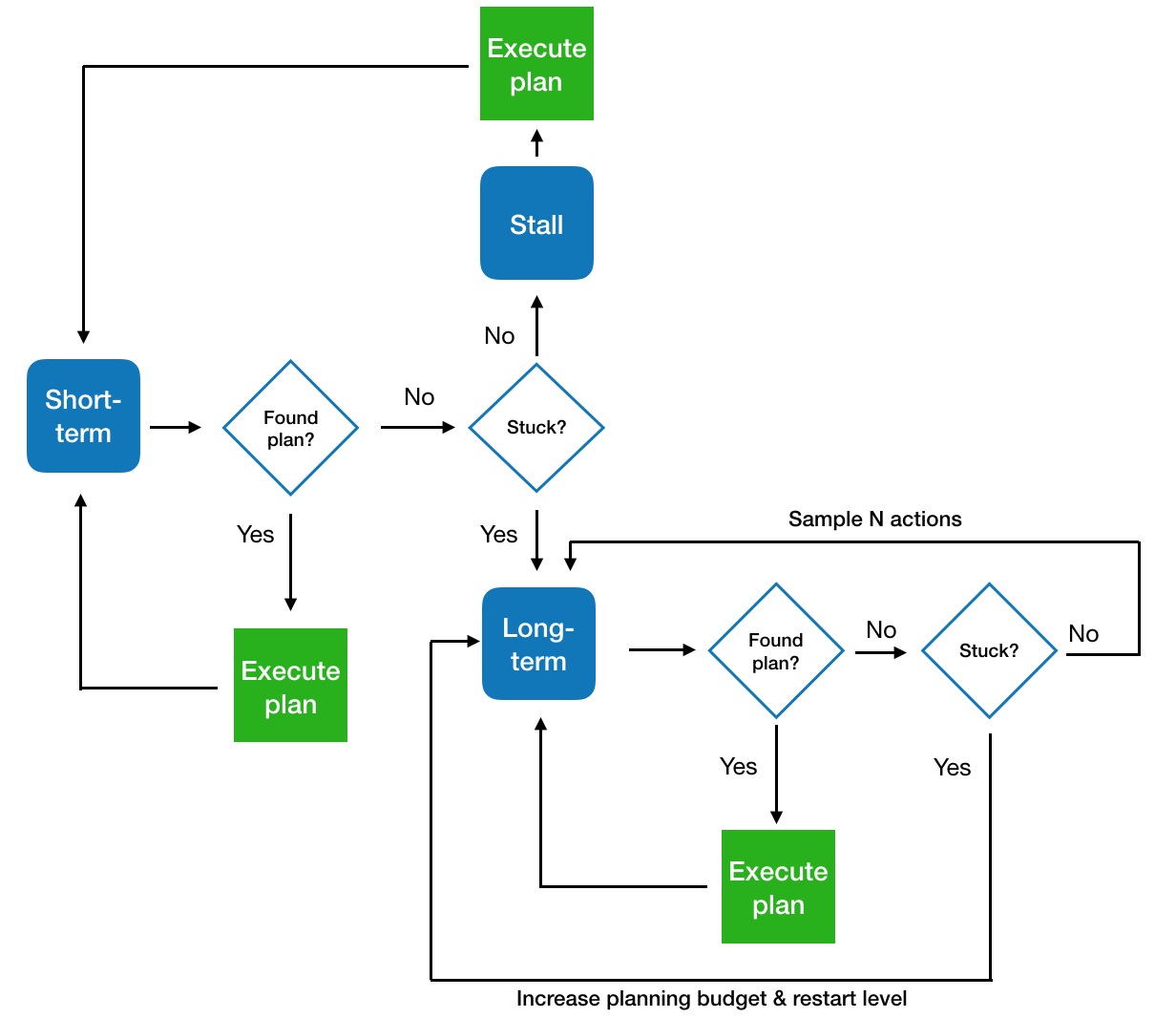}
\caption*{
{\bf Extended Data Figure 9:}
The metacontroller determines which of EMPA's three planning modes are engaged at a given moment of gameplay. The planner always begins a new game in short-term mode, and switches to other modes when it fails to find a plan within the current search budget. Which mode it switches to depends on whether it considers itself ``stuck''. The agent considers itself stuck in short-term mode when it has either lost twice in the same way, or the game contains no moving objects other than the agent itself (and hence no new possibilities to act constructively will arise merely by waiting); in each of these cases the planner switches to long-term mode. If the planner fails but the agent is not stuck, it switches to stall mode for one planning cycle. %
The agent may also consider itself stuck in long-term mode, if it has failed to find a plan from a given state more than once, %
in which case it restarts the level and doubles its planning budget. %
}
\end{figure}

\bibliography{main}

\clearpage

\end{document}